\documentclass[final]{cvpr}

\usepackage{bbm}
\usepackage{times}
\usepackage{epsfig}
\usepackage{graphicx}
\usepackage{amsmath}
\usepackage{amssymb}
\usepackage{multicol}
\usepackage{multirow}
\usepackage{booktabs}
\usepackage{enumitem}
\usepackage{pifont}
\usepackage{algorithm,algorithmic}
\usepackage{float}
\usepackage[dvipsnames]{xcolor}

\usepackage[pagebackref=true,breaklinks=true,colorlinks,bookmarks=false]{hyperref}

\begin{document}

\title{Learning to Predict Visual Attributes in the Wild}

\author{Khoi~Pham$^{1}$\thanks{A portion of this work was done during Khoi Pham's internship at Adobe Research.} \qquad Kushal~Kafle$^2$ \qquad Zhe~Lin$^2$ \qquad Zhihong~Ding$^2$ \qquad Scott~Cohen$^2$\\ \qquad Quan~Tran$^2$ \qquad Abhinav~Shrivastava$^1$\\
  \hfill$^1$University of Maryland, College Park\hfill
  $^2$Adobe Research\hfill\mbox{ }\\
{\tt\small $^1$\{khoi,abhinav\}@cs.umd.edu \qquad  $^2$\{kkafle, zlin, zhding, scohen, qtran\}@adobe.com}}

\maketitle
\begin{abstract}

Visual attributes constitute a large portion of information contained in a scene. Objects can be described using a wide variety of attributes which portray their visual appearance (color, texture), geometry (shape, size, posture), and other intrinsic properties (state, action). Existing work is mostly limited to study of attribute prediction in specific domains. In this paper, we introduce a large-scale in-the-wild visual attribute prediction dataset consisting of over 927K attribute annotations for over 260K object instances. Formally, object attribute prediction is a multi-label classification problem where all attributes that apply to an object must be predicted. Our dataset poses significant challenges to existing methods due to large number of attributes, label sparsity, data imbalance, and object occlusion. To this end, we propose several techniques that systematically tackle these challenges, including a base model that utilizes both low- and high-level CNN features with multi-hop attention, reweighting and resampling techniques, a novel negative label expansion scheme, and a novel supervised attribute-aware contrastive learning algorithm. Using these techniques, we achieve near 3.7 mAP and 5.7 overall F1 points improvement over the current state of the art. Further details about the VAW dataset can be found at \url{http://vawdataset.com/}.

\end{abstract}

\section{Introduction}

Learning to predict visual attributes of objects is one of the most important problems in computer vision and image understanding. Grounding objects with their correct visual attribute plays a central role in a variety of computer vision tasks, such as image retrieval and search \cite{johnson2015image}, tagging, referring expression \cite{kazemzadeh2014referitgame}, visual question answering (VQA) \cite{antol2015vqa,kafle2016review}, and image captioning \cite{bernardi2016automatic}.

\begin{figure}[t]
\centering
\includegraphics[width=\linewidth]{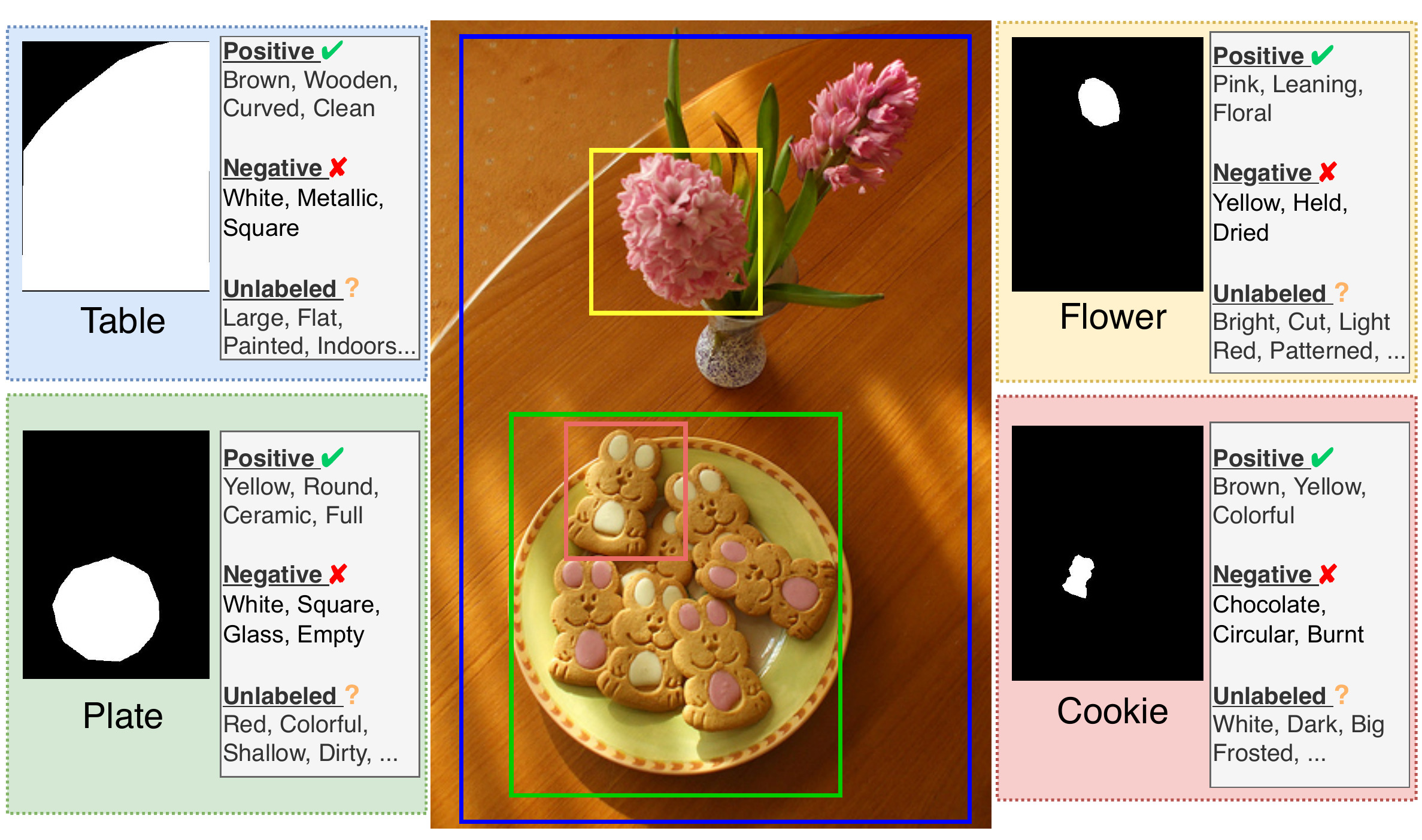}
    \caption{\textbf{Example annotations in our dataset.} Each possible attribute-object category pair is annotated with at least 50 examples consisting of explicit positive and negative labels.}
\label{fig:short}
\end{figure}

While several existing works address attribute prediction, they are limited in many ways. Objects in a visual scene can be described using a vast number of attributes, many of which can exist independently of each other. Due to the variety of possible object and attribute combinations, it is a daunting task to curate a large-scale visual attribute prediction dataset. Existing works have largely ignored large-scale visual attribute prediction in-the-wild and have instead focused only on domain-specific attributes \cite{guo2019imaterialist,li2016human}, datasets consisting of very small number of attribute-object pairs \cite{patterson2016coco}, or are rife with label noise, ambiguity and label sparsity \cite{krishna2017visual}. Similarly, while attributes can form an important part of related tasks such as VQA, captioning, referring expression, these works do not address the unique challenges of attribute prediction. Existing work also fails to address the issue of partial labels, where only a small subset of all possible attributes are annotated. Partial labels and the lack of explicit negative labels make it challenging to train or evaluate models for large-scale attribute prediction. To address these problems, we propose a new large-scale visual attribute prediction dataset for images in the wild that includes both positive and negative annotations.

Our dataset, called visual attributes in the wild (VAW), consists of over 927K explicitly labeled positive and negative attribute annotations applied to over 260K object instances (with 620 unique attributes and 2,260 unique object phrases). Due to the number of combinations possible, it is prohibitively expensive to collect exhaustive attribute annotations for each instance. However, we ensure that every attribute-object phrase pair in the dataset has a \textit{minimum} of 50 positive and negative annotations. With density of 3.56 annotations per instance, our dataset is 4.9 times denser compared to Visual Genome while also providing negative labels. Additionally, annotations in VAW are visually-grounded with segmentation masks available for 92\% of the instances. Formally, our VAW dataset proposes attribute prediction as a long-tailed, partially-labeled, multi-label classification problem.

We explore various state-of-the-art methods in attribute prediction and multi-label learning and show that the VAW dataset poses significant challenges to existing work. To this end, we first propose a strong baseline model that considers both low- and high-level features to address the heterogeneity in features required for different classes of attributes (\eg, \textit{color} vs. \textit{action}), and is modeled with multi-hop attention and an ability to localize the region of the object of interest by using partially available segmentation masks. 

We also propose a series of techniques that are uniquely suited for our problem. Firstly, we explore existing works that address label imbalance between positive and negative labels. Next, we describe a simple yet powerful scheme that exploits linguistic knowledge to expand the number of negative labels $15$-fold. Finally, we propose a supervised contrastive learning approach that allows our model to learn more attribute discriminative features. Through extensive ablations, we show that most of our proposed techniques are model-agnostic, producing improvements not only on our baseline but also other methods. Our final model is called Supervised Contrastive learning with Negative-label Expansion (SCoNE), which surpasses state-of-the-art models by 3.5 mAP and 5.7 overall F1 points.

Our paper makes the following major contributions:
\vspace{-3pt}
\begin{itemize}[noitemsep,left=0pt]
    \item We create a new large-scale dataset for visual attribute prediction in the wild (VAW) that addresses many shortcomings in existing literature and demonstrate that VAW poses considerable difficulty to existing algorithms.
    \item We design a strong baseline model for attribute prediction using existing visual attention technique. We further extend this baseline to our novel attribute learning paradigm called Supervised Contrastive learning with Negative-label Expansion (SCoNE) that considerably advances the state of the art.
    \item Through extensive experimentation, we show the efficacy of both our proposed model and our proposed techniques.
\end{itemize}

\section{Related Work}

Some of the earliest work related to attribute learning stem from a desire to learn to describe objects rather than predicting their identities \cite{farhadi2009describing, farhadi2010attribute, berg2010automatic, ferrari2008learning}. Since then, extensive work has sought to explore several aspects of object attributes, including attribute-based zero-shot object classification \cite{lampert2009learning, jayaraman2014zero, al2016recovering}, relative attribute comparison \cite{parikh2011relative, souri2016deep, prominentdifferences}, and image search \cite{siddiquie2011image, kumar2008facetracer}. While research in compositional zero shot learning \cite{wang2019tafe,purushwalkam2019task,nagarajan2018attributes,isola2015discovering} also tackle object attributes, they target transformation of `states' of objects, treat each instance as having only \textit{one} state, and focus on predicting unseen compositions rather than the prediction of complete set of attributes for each object instance. Several works have focused on attribute learning in specific domains such as animals, scenes, clothing, pedestrian, human facial and emotion attributes \cite{xian2018zero, zhou2017places, guo2019imaterialist, liu2016deepfashion, li2016human, kosti2017emotion}. In contrast, we seek to explore attribute prediction for \textit{unconstrained} set of objects. 

Only a limited number of work have sought to explore \textit{general} attribute prediction. COCO Attributes \cite{patterson2016coco} is an attempt to develop in-the-wild attribute prediction dataset, however, it is very limited in scope, covering only 29 object categories. Similarly, a portion of the the Visual Genome (VG) \cite{krishna2017visual} dataset consists of attribute annotations. However, attributes in VG are not a central focus of the work and therefore they are very sparsely labeled, noisy, and lack negative labels, making it unsuitable to be used as a standalone attribute prediction benchmark. Despite this, attribute annotations from VG are used to train attribute-aware object detectors for downstream vision-language tasks \cite{anderson2018bottom,jiang2020defense,zhang2021vinvl}. By introducing the VAW dataset, the research community can use its dense attribute annotations in conjunction with VG and our attribute learning techniques to train better attribute prediction models. Several recent works have also sought to take advantage of massive amount of data in VG to curate datasets for specific challenges \cite{hudson2019gqa, wu2020phrasecut}. In a similar vein, we also start by leveraging existing sources of clean annotations to develop our VAW dataset. 

VAW can be cast as a multi-label classification problem which has been extensively studied in the research community \cite{durand2019learning, chen2019multi, li2017improving, sarafianos2018deep}. Multi-label learning involving missing labels poses a greater challenge, but is also extensively studied \cite{huynh2020shared,kundu2020exploiting,wu2018multi,yang2016improving}. In many cases, missing labels are assumed to be \textit{negative} examples \cite{sun2010multi,bucak2011multi,sun2017revisiting,yu2014large} which is unsuitable for attribute prediction, since most of the attributes are not mutually exclusive. Some others attempt to predict missing labels by training expert models \cite{shin2020semi}, which is also infeasible for a large-scale problem like ours.

Data imbalance naturally arises in datasets with large set of labels. As expected, label imbalance exists in our VAW dataset, therefore techniques designed to learn from imbalanced data are also related to our explorations. These works can be divided into two main approaches: cost-sensitive learning \cite{huang2016learning,wang2017learning,cui2019class} and resampling \cite{shen2016relay,he2009learning,buda2018systematic,gupta2019lvis,mahajan2018exploring}. We utilize both of these techniques in our final model.

Attention is a highly effective technique in image classification, captioning, VQA, and domain-specific attribute prediction \cite{wang2017multi,xu2015show,lu2016hierarchical,huynh2020shared,tay2019aanet,zhang2019task,sarafianos2018deep}. In our VAW dataset, most of the objects are annotated with their segmentation mask, which allows us to guide the attention map to ignore irrelevant image regions. We also use additional attention maps to allow our model to properly explore the surrounding context of the object.

Contrastive learning has recently gained a lot of traction as an effective self-supervised learning technique \cite{wu2018unsupervised,bachman2019learning,he2020momentum,chen2020simple}. While originally intended to be used in self-supervised setting, recent works have expanded contrastive learning to be used in supervised setting \cite{khosla2020supervised}. Motivated by these works, we propose an extension of supervised contrastive loss to allow it to work in a multi-label setting required for VAW. To the best of our knowledge, ours is the first attempt to apply contrastive loss for multi-label learning.

\section{Visual Attributes in the Wild (VAW) Dataset}
In this section, we describe how we collect attribute annotations and present statistics of the final VAW dataset. In general, we aim to overcome the limitations of VG on the attribute prediction task which includes noisy labels, label sparsity, and lack of negative labels to create a dataset applicable for training and testing attribute classification models.

\vspace{-2pt}
\subsection{Data collection}

VAW is created based on the VGPhraseCut and GQA datasets, both of which leverage and refine annotations from Visual Genome. VGPhraseCut is a referring expression dataset that provides high-quality attribute labels and per-instance segmentation mask, while GQA is a VQA dataset that presents cleaner scene graph annotations.\smallskip

\noindent \textbf{Step 1: Extraction from VGPhraseCut and GQA}

Our goal is to build a dataset that allows us to predict the maximal number of attributes \textit{commonly} used to describe objects in the wild. From VGPhraseCut, we select attributes that appear within more than 15 referring phrases. After manually cleaning ambiguous/hard to recognize attributes, we obtain a set of 620 unique attributes which are used throughout the rest of the process. Next, we extract more instances from GQA that are labeled with these attributes. We further take advantage of the referring expressions from VGPhraseCut to collect a reliable negative label set: given an image, for instances that are not selected by an attribute referring phrase, we assign that attribute as a negative label for the instance. This step allows us to collect 220,049 positive and 21,799 negative labels.

\vspace{2pt}
\noindent \textbf{Step 2: Expand attribute-object coverage}

In this step, we seek to collect additional annotations for every \textit{feasible} attribute-object pair that may be lacking annotations. We define feasible pair as those with at least 1 positive example in our dataset. We ensure that every feasible pair has at least 50 (positive or negative) annotations. To keep the annotation cost in check, we do not annotate pairs that already have 50 or more annotations. This expansion enriches our dataset with more positives and negatives for every attribute across different objects, allowing for better training and evaluation of classification models. This step adds 156,690 positive and 455,151 negative annotations.\smallskip

\vspace{1pt}
\noindent \textbf{Step 3: Expand long-tailed attribute set}

In this step, we aim to collect additional annotations for the long-tailed attributes. Long-tailed attributes are associated with very few object categories, which is either due to the attribute not being frequently used by humans or the attribute is only applied to a small set of objects. Hence, given a long-tailed attribute and a known object that it applies to, we first expand its set of \textit{possibly applied objects} by using the WordNet ontology. For example, while \textit{playing} may only be applied to \textit{child} in the training set, it \textit{could} also be applicable to other closely related object categories like \textit{man, woman, person}. After we find candidate object categories for a given long-tail attribute, we ask humans to annotate randomly sampled images from these candidates with either positive or negative label for the given attribute. This step adds additionally 16,239 positive and 57,751 negative annotations pertaining to all long-tailed attributes. 

\subsection{Statistics}

Our final dataset consists of 620 attributes describing 260,895 instances from 72,274 images. Our attribute set is diverse across different categories, including \textit{color, material, shape, size, texture, action}. On the annotated instances, our dataset consists of 392,978 positive and 534,701 negative attribute labels. The instances from VGPhraseCut (occupy 92\% in the dataset) are provided with segmentation masks which can be useful in attribute prediction. We split the dataset into 216,790 instances (58,565 images) for training, 12,286 instances (3,317 images) for validation, and 31,819 instances (10,392 images) for testing. We split the dataset such that the test set has higher annotation density per object, which allows for more thorough testing. In particular, our test set has an average of 7.03 annotations per instance compared to 3.02 in the training set.

In Table \ref{table:comparison}, we compare the statistics of the VAW dataset with other in-the-wild and domain-specific visual attribute datasets. Compared to existing work, VAW fills an important gap in the literature by providing a domain-agnostic, in-the-wild visual attribute prediction dataset with denser annotations, explicit negative labels, segmentation masks, and large number of attribute and object categories.

\begin{table*}
\resizebox{\textwidth}{!}{
\begin{tabular}{@{}llllllll@{}}
    \toprule
    \textbf{Dataset} & \textbf{VAW} & \textbf{Visual Genome} \cite{krishna2017visual} & \textbf{COCO Attributes} \cite{patterson2016coco} & \textbf{EMOTIC} \cite{kosti2017emotion} & \textbf{WIDER} \cite{li2016human} & \textbf{iMaterialist} \cite{guo2019imaterialist} \\
    \midrule
    \# attributes & 620 & 68,111 & 196 & 26 & 14 & 228 \\
    \# instances & 260,895 &  3,843,636 & 180,000 & 23,788 & 57,524 & 1,012,947 \\
    \# object categories & 2,260 & 33,877 & 29 & 1 (person)* & 1 (person*) & 1 (clothes)* \\
    \# attribute anno. per instance & 3.56 & 0.73 & $\geq$ 20 & 26 & 14 & 16.17 \\
    
    Negative Labels & Yes & No & Yes & Yes & Yes & Yes \\
    Segmentation masks & Yes & No & No & No & No & Yes \\
    Domain & In-the-wild & In-the-wild & In-the-wild & Emotions & Pedestrian & Fashion \\
    \bottomrule
\end{tabular}
}
\\
\caption {\textbf{Statistics of VAW with other attribute in-the-wild and domain-specific datasets.} *person (resp. *clothes) category may represent multiple categories including \{boy, girl, man, woman, \etc\} (resp. \{shirt, pants, top, \etc\}). While Visual Genome is the largest among these in terms of number of attribute annotations, it is sparsely labeled. Other datasets are either fully annotated for domain-specific attributes or more densely labeled but covering few object categories. }
 \label{table:comparison}
\vspace{0mm}
\end{table*}

\section{Methodology}

\begin{figure*}
\centering
\includegraphics[width=0.90\linewidth]{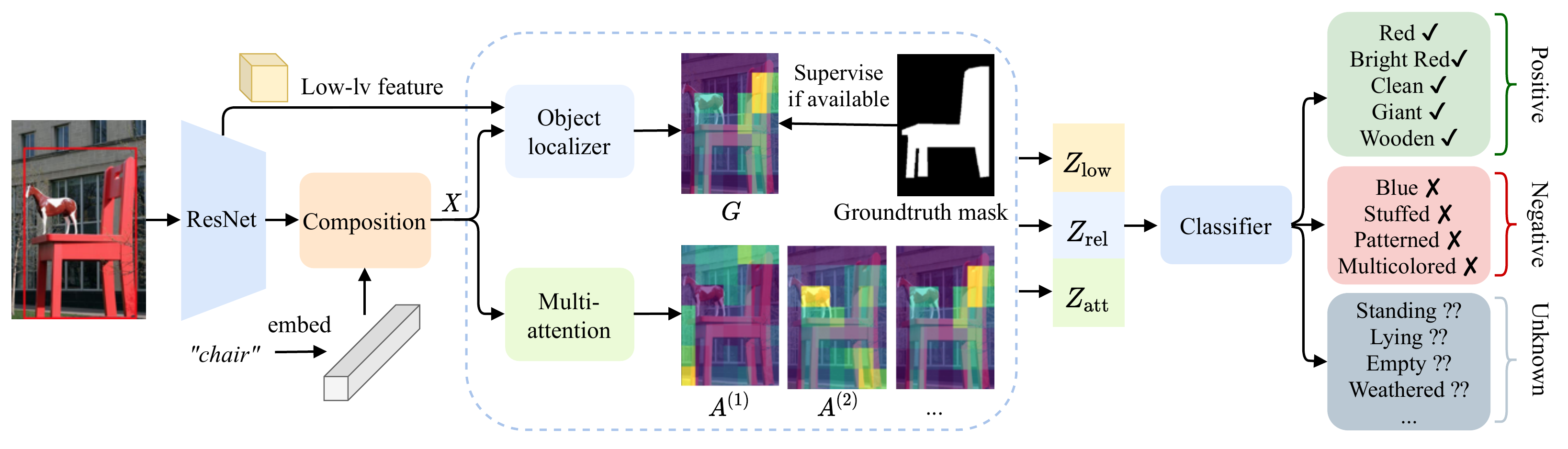}
    \caption{\textbf{Strong baseline attribute prediction model.} Feature map extracted from the input image is modulated with the object embedding which allows the model to learn useful attribute-object relationships (\eg \textit{ball} is \textit{round}) and also to suppress infeasible attribute-object pairs (\eg \textit{talking table}). The image-object combined feature map $X$ is used to infer the object region $G$ and multiple attention maps $\{A^{(m)}\}$ which are subsequently used to aggregate features for classification. Here, $Z_\text{low}$ and $Z_\text{rel}$ respectively denotes low-level and image-object features aggregated inside the estimated object region, $Z_\text{att}$ corresponds to image-object features pooled from the multiple attention maps. The classifier is trained with BCE loss on the explicit positive and negative labels. For the missing (unknown) labels, we find treating them as ``soft negatives'' and assigning them with very small weights in the BCE loss also helps improve results.} 
\label{fig:model}
\end{figure*}

In this section, we will describe components of our strong baseline model along with the Supervised Contrastive learning with Negative-label Expansion (SCoNE) algorithm that helps our model to learn more attribute discriminative features. A depiction of our strong baseline model is shown in Figure \ref{fig:model}.

\subsection{Preliminaries}\label{sec:prelim}

\noindent \textbf{Problem formulation: } Let $\mathcal{D} = \{I_i,g_i,o_i;Y_i\}_{i=1}^{N}$ be a dataset of $N$ training samples, where $I_i$ is an object instance image (cropped using its bounding box), $g_i$ is its segmentation mask, $o_i$ is the category phrase of the object for which we want to predict attributes, and $Y_i = [y_{i,1}, ..., y_{i,C}]$ is its $C$-class label vector with $y_c \in \{1, 0, -1\}$ denoting whether attribute $c$ is positive, negative, or missing respectively. Our goal is to train a multi-label classifier that, given an input image and the object name, can output the confidence score for all $C$ attribute labels.

\noindent \textbf{Image feature representation:} Given an image $I$ of an object $o$, let $f_\text{img}(I) \in \mathbb{R}^{H\times W \times D}$ be the $D$-dimensional image feature map with spatial size $H \times W$ extracted using any CNN backbone architecture. In our model, we use the output of the penultimate layer of ResNet-50 \cite{he2016deep}.

\subsection{Image-object feature composition}
\label{sec:img_obj_comp}
Prior models for attribute prediction mostly tackle domain-specific settings or a limited number of object categories \cite{patterson2016coco,sarafianos2018deep,shin2020semi}. Hence, these works are able to employ object-agnostic attribute classification. However, because our VAW dataset contains attribute annotations across a diverse set of object categories, incorporating object embedding as input can help the model learn to avoid infeasible attribute-object combinations (\eg \textit{parked dog}).

There are multiple ways to compose the image feature map with the object embedding \cite{perez2018film,vo2019composing,anderson2018bottom}. Here, we opt for a simple object-conditioned gating mechanism, which we find to be consistently better than concatenation used in \cite{anderson2018bottom,jiang2020defense}. Let $\phi_o \in \mathbb{R}^d$ be the object embedding vector, $f_\text{comp}(f_\text{img}(I), \phi_o) \in \mathbb{R}^D$ be the composition module that takes in the image feature map and object embedding. We implement $f_\text{comp}$ with a gating mechanism as follows:
\begin{equation}
    f_\text{comp}(f_\text{img}(I), \phi_o) = f_\text{img}(I) \odot f_\text{gate}(\phi_o),
\end{equation}
\begin{equation}
    f_\text{gate}(\phi_o) = \sigma(W_{\text{g}_2} \cdot  \text{ReLU}(W_{\text{g}_1}\phi_o + b_{\text{g}_1}) + b_{\text{g}_2}),
\end{equation}
$\odot$ is the channel-wise product, $\sigma(\cdot)$ is the sigmoid function, $f_\text{gate}(\phi_o) \in \mathbb{R}^D$ is broadcasted to match the feature map spatial dimension and is a 2-layer MLP. Intuitively, $f_\text{gate}$ acts as a filter that only selects attribute features relevant to the object of interest and suppresses incompatible attribute-object pairs.

\subsection{Object localization and multi-attention module}

\noindent \textbf{Relevant object localizer.} An object bounding box can contain both the relevant object and other objects or background. Hence, it is desirable to learn a smarter feature aggregation that can suppress all irrelevant image regions. We propose to leverage the availability of the object segmentation mask in the VAW dataset to achieve this.

Let $X \in \mathbb{R}^{H\times W \times D}$ be the image-object composed feature map, the relevant object region $G$ is localized using a 2-stacked convolutional layers $f_\text{rel}$ with kernel size 1, followed by spatial softmax:
\begin{align}
    &g = f_\text{rel}(X), \,\,\, g \in \mathbb{R}^{H \times W}, \\
    G_{h,w} &= \frac{\exp(g_{h,w})}{\sum_{h,w}\exp(g_{h,w})}, \,\,\, G \in \mathbb{R}^{H \times W}.
\end{align}

We can then pool the image feature vector as
\begin{equation}
    Z_\text{rel} = \sum_{h,w}G_{h,w}X_{h,w}.
\end{equation}
$G$ is learned with direct supervision from the object mask whenever available with the following loss:

\begin{equation}
    \mathcal{L}_\text{rel} = \sum_{h,w}(G_{h,w} \times (1-M_{h,w})) - \lambda_\text{rel}(G_{h,w} \times M_{h,w}),
\end{equation}
where $M$ is the ground truth object binary mask.
Rather than requiring $G$ to exactly match the object mask, we find it is better to penalize the network whenever its prediction falls outside of the mask. This frees the network to learn heterogeneous attention within the object region if necessary (\eg, \textit{black mirror} refers to its frame being \textit{black} rather than its interior) instead of distributing its attention uniformly over the object. Hence, by setting $\lambda_\text{rel}$ to a small positive constant less than $1$, we prioritize the need for $G$ to \textit{not} attend to non-object pixels over the need to uniformly attend to object pixels.

\noindent \textbf{Multi-attention module.} \,\,Object localization is beneficial for recognizing several attributes such as \textit{color, material, texture,} and \textit{shape}, but might be too restrictive for attributes that require attention to different object parts or the background. For example: \textit{bald-headed} or \textit{bare-footed} requires looking at a person's head or foot; distinguishing different activities (e.g., \textit{jumping} vs \textit{crouching}) might require information from surrounding context. Therefore, we utilize a free-form multi-attention mechanism to allow our algorithm to attend to features at different spatial locations.

There are two extreme cases to apply spatial attention \cite{huynh2020shared}: (1) one attention map for all attributes and (2) one attention map per attribute \cite{sarafianos2018deep}. The first approach is similar to using the object foreground which is unlike what we are aiming for. The latter allows for more control but does not scale well with large number of attributes. Hence, we opt for a hybrid multi-attention idea as in \cite{huynh2020shared}.

We extract $M$ attention maps $\{A^{(m)}\}_{m=1}^{M}$ from $X$ using $f_\text{att}^{(m)}$ which has the same architecture as $f_\text{rel}$:
\begin{align}
    E^{(m)} &= f_\text{att}^{(m)}(X), \,\,\, E^{(m)} \in \mathbb{R}^{H \times W}, m = 1,...,M\\
    A_{h,w}^{(m)} &= \frac{\exp(E_{h,w}^{(m)})}{\sum_{h,w}\exp(E_{h,w}^{(m)})}, \,\,\, A^{(m)}_{h,w} \in \mathbb{R}^{H \times W}.
\end{align}

This is partly similar to \cite{zhao2019recognizing} where object parts are localized using learned embeddings of these parts. Because the VAW dataset does not have part annotations for every attribute, this approach is not usable in our case. Similar to \cite{huynh2020shared}, we employ the following divergence loss to encourage these attention maps to focus on different regions:
\begin{equation}
    \mathcal{L}_\text{div} = \sum_{m \neq n}\frac{\langle E^{(m)},E^{(n)}\rangle}{\|E^{(m)}\|_2\|E^{(n)}\|_2}.
\end{equation}

Using the computed $M$ attention maps, we aggregate $M$ feature vectors $\{r^{(m)}\}_{m=1}^{M}$ from $X$ and pass them through a projection layer to obtain their final representations:
\begin{equation}
    r^{(m)} = \sum_{h,w}A_{h,w}^{(m)}X_{h,w}, \,\,\,r^{(m)} \in \mathbb{R}^D,
\end{equation}
\begin{equation}
    z^{(m)}_\text{att} = f_\text{proj}^{(m)}(r^{(m)}), \,\,\,z_\text{att}^{(m)} \in \mathbb{R}^{D_\text{proj}}.
\end{equation}

Our final multi-attention feature is the concatenation of all individual attention features:

\begin{equation}
    Z_\text{att} = \text{concat}([z^{(1)}_\text{att},...,z^{(M)}_\text{att}]).
\end{equation}

\subsection{Loss function and training paradigm} \label{sec:resample}
Our final feature vector is the concatenation of the localized object and the multi-attention feature. In addition, we also find using low-level feature from early blocks improves accuracy for low-level attributes (\textit{color, material}). Therefore, we also pool low-level features from the estimated object region $G$ to construct $Z_\text{low}$. The input to the classification layer is $[Z_\text{low}, Z_\text{rel}, Z_\text{att}]$, and we use a linear classifier with $C$ output logit values followed by sigmoid.

Let $\hat{Y} = [\hat{y}_{1},...,\hat{y}_{C}]$ be the output of the classification layer. We apply the following reweighted binary cross-entropy loss that takes data imbalance into account:
\begin{align}
    \mathcal{L}_\text{bce}(Y, \hat{Y}) = -\sum_{c=1}^{C} w_c\big(&\mathbbm{1}_{[y_{c}=1]} p_{c} \log(\hat{y}_{c})\\ + &\mathbbm{1}_{[y_{c}=0]} n_c\log(1-\hat{y}_{c}) \big), \nonumber
\end{align}
where $w_c, p_c, n_c$ are respectively the reweighting factors for attribute $c$, its positive, and its negative example. Let $n_c^\text{pos}$ and $n_c^\text{neg}$ be the number of positives and negatives of attribute $c$. First, we want $w_c$ to reflect the importance of the rare attributes by setting $w_c$ inversely proportional to its number of positive examples: $w_c \propto 1/(n_c^\text{pos})^\alpha$ and normalize so that $\sum_c w_c = C$ \cite{cui2019class} ($\alpha$ is a smoothing factor). Second, we want to balance between the effect of positive and negative examples. We apply the same idea by setting $p_c \propto 1/(n_c^\text{pos})^\alpha$, $n_c \propto 1/(n_c^\text{neg})^\alpha$ and normalize so that $p_c + n_c = 2$. As a result, the ratio between the positive and negative becomes $p_c/n_c = (n_c^\text{neg}/n_c^\text{pos})^\alpha$, which helps balance out their effect based on their frequency.

Our re-weighted BCE (termed RW-BCE) is different from \cite{durand2019learning}, where the authors propose to reweigh each sample based on its proportion of available labels (\ie, an instance with less number of available labels is assigned a larger weight). We posit this is not ideal because the number of labels for an instance should not affect loss computation (\eg, loss for \textit{red} should be the same between a \textit{red car} instance and a \textit{large shiny red car} instance). Our final loss is a combination of all loss functions presented above:
\begin{equation}
    \mathcal{L} = \mathcal{L}_\text{bce} + \mathcal{L}_\text{rel} + \lambda_\text{div}\mathcal{L}_\text{div}.
\end{equation}

Empirically, we find applying repeat factor sampling (RFS) \cite{gupta2019lvis,mahajan2018exploring}  with RW-BCE works well. RFS is a method that defines a repeat factor for every image based on the rarity of the labels it contains. Therefore, we employ both RW-BCE and RFS (referred as RR) in training our model.

\subsection{Negative label expansion}

While our dataset provides unprecedented amount of explicitly labeled negative annotations, the amount of possible negatives still far outnumbers the number of possible positive attributes. Because many attributes are mutually exclusive (\ie, presence of attribute \textit{clean} implies absence of attribute \textit{dirty}), we seek to use existing linguistic and external knowledge tools to expand the set of negative annotations.

Consider attribute type $A$ (\eg, \textit{material}), the following observations can be made about its attributes: (1) there exists \textit{overlapping} relation between some attributes due to their visual similarity or them being hierarchically related (\eg, \textit{wooden} overlaps with \textit{wicker}); (2) there exists \textit{exclusive} relation where two attributes cannot appear on the same object (\eg, \textit{wet} vs. \textit{dry}). Therefore, for an object labeled with attribute $a \in A$, we can generate negative labels for it from the set $\{a' \in A \mid \neg \text{overlap}(a, a') \vee \text{exclusive}(a, a')\}$.

We classify the attributes into types and construct their overlapping and exclusive relations using existing ontology from a related work \cite{Han2019Visual}, WordNet \cite{wordnet}, and the relation edges from ConceptNetAPI \cite{speer2016conceptnet}. We further expand the overlapping relations based on the co-occurrence (by using conditional probability) of the attribute pairs (\eg, \textit{white} and \textit{beige} are similar and often mistaken by human annotators). More details about how these relations are constructed are presented in the supplementary material. Our negative label expansion scheme allows to add 5.9M negative annotations to our training set. Aside from the extra negatives, one strong point of this approach is that when we want to label a novel attribute class, we can use the same approach to discover its relationship with existing attributes in the dataset and attain free negatives for the new class.

\subsection{Supervised contrastive learning}
\cite{yang2020rethinking} shows with success that imbalanced learning could benefit from self-supervised pretraining on both labeled and unlabeled data, where a network can be better initialized by originally avoiding strong label bias due to data imbalance. Also motivated by \cite{khosla2020supervised}, we propose to use supervised contrastive (SupCon) pretraining for our attribute learning with partial labels problem, where we extend the SupCon loss from a single-label to a multi-label setting.

We perform mean-pooling inside the feature map $X$ to obtain $x \in \mathbb{R}^{D}$. We follow the design of SimCLR \cite{chen2020simple} and add a projection layer to map $z = \text{Proj}(x) \in \mathbb{R}^{128}$. The projection layer is an MLP with hidden size 2048 and is only used during pretraining. In a multi-label setting, it is not trivial how to pull two samples together since they can share some labels but different in terms of other labels. Motivated by \cite{nagarajan2018attributes,hewitt2019structural}, we propose to represent each attribute $c$ as a matrix $A_c \in \mathbb{R}^{128 \times 128}$ that linearly projects $z$ into an attribute-aware embedding space $z_{c} = A_cz$, which is then $\ell_2$-normalized onto the unit hypersphere. With this, samples that share the same attribute can have their respective attribute-aware embeddings pulled together.

In the pretraining stage, we construct a batch of $2N$ sample-label vector pairs $\{I_i,Y_i\}_{i=1}^{2N}$, where $I_{2k}$ and $I_{2k-1}$ ($k=1...N$) are two views (from random augmentation) of the same object image and $Y_{2k}=Y_{2k-1}$. Let $z_{i,c}$ be the $c$-attribute-aware embedding of $I_i$, and $B(i) = \{c \in C : Y_{i,c}=1 \}$ is the set of positive attributes of $I_i$.
We reuse notations from \cite{khosla2020supervised}: $K \equiv \{1...2N\}$, $A(i) \equiv K \setminus \{i\}$, $P(i,c) \equiv \{p \in A(i): Y_{p,c} = Y_{i,c}\}$ 
and use the following SupCon loss
\begin{equation}
    \mathcal{L}_\text{sup} = \sum_{i=1}^{2N} \sum_{c \in B(i)} \frac{-1}{|P(i,c)|}\sum_{p\in P(i,c)}\log\frac{\exp{(z_{i,c} \cdot z_{p,c}/\tau)}}{\sum\limits_{j \in A(i)}\exp(z_{i,c} \cdot z_{j,c}/\tau)}.
\end{equation}

Linear transformation using $A_c$, followed by dot product in the SupCon loss, implements an inner product in the embedding space of $z$, which can be interpreted as finding part of $z$ that encodes the attribute $c$ \cite{hewitt2019structural}. Therefore, our approach fits nicely into the multi-label setting where an image embedding vector $z$ can simultaneously encode multiple attribute labels that can be probed by these linear transformations for contrasting in the SupCon loss.

After the pretraining stage, we keep the backbone encoder and the image-object composition module and finetune them along with the classification layer.

While SupCon is designed to be used for pretraining, empirically, we find it hampers the multi-attention module ability to focus on specific regions. To reconcile this difference, we find it is empirically better to minimize $\mathcal{L}_\text{sup}$ jointly with the other loss.
For other models that do not use attention (vanilla ResNet), we find SupCon pretraining still effective.

\section{Experiments}

\begin{table*}[t]
\resizebox{\textwidth}{!}{
\centering
\begin{tabular}{@{}lcccccccccccccc@{}}
    \toprule
    \multirow{2}{*}{\textbf{Methods}} &
    \multicolumn{4}{c}{\textbf{Overall}} & \multicolumn{3}{c}{\textbf{Class imbalance (mAP)}} & \multicolumn{7}{c}{\textbf{Attribute types (mAP)}} \\
    \cmidrule(r){2-5} \cmidrule(lr){6-8} \cmidrule(l){9-15}
  & \textbf{mAP} & \textbf{mR@15} & \textbf{mA} & \textbf{F1@15} & \textbf{Head} & \textbf{Medium} & \textbf{Tail} & \textbf{Color} & \textbf{Material} & \textbf{Shape} & \textbf{Size} & \textbf{Texture} & \textbf{Action} & \textbf{Others} \\
    \midrule
    LSEP \cite{li2017improving} & 61.0 & 50.7 & 67.1 & 62.3 & 69.1 & 57.3 & 40.9 & 56.1 & 67.1 & 63.1 & 61.4 & 58.7 & 50.7 & 64.9 \\
    ML-GCN \cite{chen2019multi} & 63.0 & 52.8 & 69.5 & 64.1 & 70.8 & 59.8 & 42.7 & 59.1 & 64.7 & 65.2 & 64.2 & 62.8 & 54.7 & 66.5 \\
    Partial-BCE + GNN \cite{durand2019learning} & 62.3 & 52.3 & 68.9 & 63.9 & 70.1 & 58.7 & 40.1 & 57.7 & 66.5 & 64.1 & 65.1 & 59.3 & 54.4 & 65.9 \\
    \midrule
    ResNet-Baseline \cite{patterson2016coco} & 63.0 & 52.1 & 68.6 & 63.9 & 71.1 & 59.4 & 43.0 & 58.5 & 66.3 & 65.0 & 64.5 & 63.1 & 53.1 & 66.7 \\
    ResNet-Baseline-CE \cite{anderson2018bottom,jiang2020defense} & 56.4 & 55.8 & 50.3 & 61.5 & 64.6 & 52.7 & 35.9 & 54.0 & 64.6 & 55.9 & 56.9 & 54.6 & 47.5 & 59.2 \\
    Sarafianos \etal \cite{sarafianos2018deep} & 64.6 & 51.1 & 68.3 & 64.6 & 72.5 & 61.5 & 42.9 & 62.9 & 68.8 & 64.9 & 65.7 & 62.3 & 56.6 & 67.4 \\
    Strong Baseline (SB) & 65.9 & 52.9 & 69.5 & 65.3 & 73.6 & 62.5 & 46.0 & 64.5 & 68.9 & 67.1 & 65.7 & 66.1 & 57.2 & 68.7 \\
    \midrule
    \textbf{SB + SCoNE (Ours)} & \textbf{68.3} & \textbf{58.3} & \textbf{71.5} & \textbf{70.3} & \textbf{76.5} & \textbf{64.8} & \textbf{48.0} & \textbf{70.4} & \textbf{75.6} & \textbf{68.3} & \textbf{69.4} & \textbf{68.4} & \textbf{60.7} & \textbf{69.5} \\
    \bottomrule
\end{tabular}
}
\\
\caption {\textbf{Experimental results compared with baselines and SOTA multi-label learning methods.} The top box displays results of multi-label learning methods; the middle box shows results of models from attribute prediction works and our strong baseline; the last row shows performance of our SCoNE algorithm applied onto the strong baseline.
}
\label{table:overall}
\end{table*}

\begin{table} 
\footnotesize
\begin{tabular}{@{}lcccc@{}}
    \toprule
    \textbf{Methods} &
    \textbf{mAP} & \textbf{mR@15}  & \textbf{mA} & \textbf{F1@15} \\
    \midrule
    Strong Baseline (SB) & 65.9 & 52.9 & 69.5 & 65.3 \\
    + Negative & 67.7 & 54.3 & 70.0 & 69.6 \\
    + Neg + SupCon & 68.2 & 55.2 & 70.3 & 70.0 \\
    + Neg + SupCon + RR (SCoNE) & \textbf{68.3} & \textbf{58.3} & \textbf{71.5} & \textbf{70.3} \\
    \midrule
    ResNet-Baseline & 63.0 & 52.1 & 68.6 & 63.9 \\
    + SCoNE & 66.4 & 56.8 & 70.7 & 68.8 \\
    \bottomrule
\end{tabular}\\
\caption {\textbf{Ablation study.} We show how each of our proposed techniques help improve overall performance.}
\label{table:ablation}
\vspace{0mm}
\end{table}

In this section, we discuss the evaluation metrics and report the results of our attribute prediction framework and other related baselines on the VAW dataset. The implementation details are presented in the supplementary material.

\subsection{Algorithms for VAW dataset}
We consider the following baselines and state-of-the-art multi-label learning approaches and compare them to our SCoNE algorithm. We made our best attempt to modify the authors' implementation (if available) to include the image-object composition module in section \ref{sec:img_obj_comp}. All models use ResNet-50 as their backbone and use BCE loss (except LSEP and ResNet-Baseline-CE) for training. Empirically, we find treating missing labels as negatives and assigning them with very small weights in the BCE loss also improves results. Hence, we apply this for all methods.
\begin{itemize}[noitemsep,left=0pt]
    \item \textbf{ResNet-Baseline:} ResNet-50  followed by image-object composition and classification layer.
    \item \textbf{ResNet-Baseline-CE:} Similar as above, but uses softmax cross entropy loss. This is used by \cite{anderson2018bottom,jiang2020defense} to train attribute prediction head for object detectors on Visual Genome.
    \item \textbf{Strong Baseline (SB)}: The combination of our image-object composition, multi-attention, and object localizer.
    \item \textbf{LSEP} \cite{li2016human} \textbf{:} Uses ranking loss and label threshold estimation method to predict which attributes to output.
    \item \textbf{ML-GCN} \cite{chen2019multi} \textbf{:} Uses graph convolution network to predict classifier weights based on the 100-d GloVe embeddings of the attribute names. Label correlation graph is constructed following the authors' implementation.
    \item \textbf{Durand \etal (Partial BCE + GNN)} \cite{durand2019learning} \textbf{:} BCE loss reweighted by the authors' reweighting scheme. Graph neural network is applied on the output logits.
    \item \textbf{Sarafianos \etal} \cite{sarafianos2018deep} \textbf{:} one of the SOTAs in pedestrian attribute prediction that also uses multi-attention.
\end{itemize}
    
\subsection{Evaluation metrics}

We employ the following metrics to measure attribute prediction from different perspectives. Detailed discussion of these metrics can be found in the supplementary.

\noindent \textbf{mAP:} mean average precision over all classes. mAP is a popular metric for attribute prediction and multi-label learning \cite{patterson2016coco, chen2019multi, veit2017learning}. Because VAW is partially labeled, we only evaluate mAP on the annotated data as in \cite{veit2017learning}.

\noindent \textbf{mR@15:} mean recall over all classes at top 15 predictions in each image. Recall@K is often used in datasets that are not exhaustively labeled such as
scene graph generation \cite{zareian2020bridging,chen2019knowledge}. This is also used in multi-label learning \cite{chen2019multi,li2017improving,huynh2020shared} under the name `per-class recall'.

\noindent \textbf{F1@15}: as the above metric may be biased towards infrequent classes, we also report overall F1 at top 15 predictions in each image. Because VAW is partially labeled, we only evaluate the prediction of label that has been annotated.

\noindent \textbf{mA:} mean balanced accuracy over all classes using threshold at 0.5 to separate between positive and negative prediction. This metric is used in pedestrian and human facial attribute works \cite{sarafianos2018deep,li2016richly}.

\subsection{Results}

\begin{figure}
\centering
\includegraphics[width=0.95\linewidth]{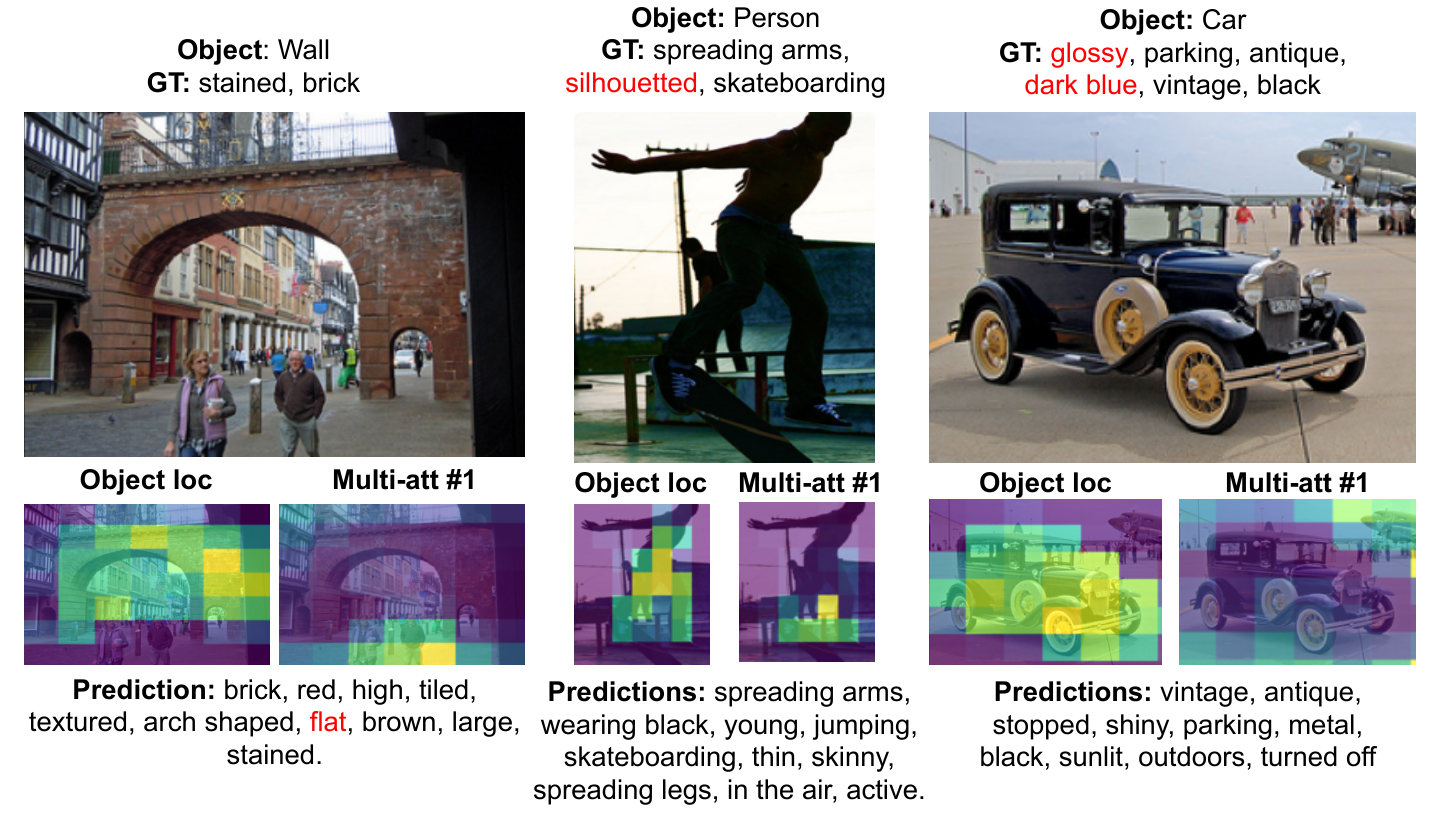}
    \caption{\textbf{Qualitative results.} Examples of predictions from SB+SCoNE. We show the object name and its ground truth positive attribute labels above the image. The object localized region, attention map \#1, and model top-10 predictions are shown below. Red text represents missed or incorrect predictions.}
\label{fig:exampls}
\end{figure}

Overall results are shown in Table \ref{table:overall}, where SB and SB+SCoNE are compared with other baselines and state-of-the-art algorithms. In overall, SB is better than other baselines in almost all metrics except for mR@15 where it is lower than ResNet-Baseline-CE. This shows that the object localizer and multi-attention are effective in attribute prediction. ResNet-Baseline-CE, which is adopted by \cite{anderson2018bottom,jiang2020defense}, has good recall but very low precision (mAP and F1). This is in contrast to ResNet-Baseline which is trained with BCE.

SB+SCoNE substantially improves over SB in all metrics and clearly surpasses available algorithms by a large margin. It is particularly effective in long-tail attributes where it outperforms its closest competitor (other than SB) by 5 mAP points, and is also highly effective in detecting color and material attributes where it is nearly 7-8 mAP points higher than the next-best method. This shows that our attribute learning paradigm, including the negative label expansion, supervised contrastive loss, reweighted and resampling scheme is clearly effective in attribute learning.

\subsection{Ablation studies} 
\label{sec:ablation}
Table \ref{table:ablation} shows effect of different components of SCoNE. Starting from our SB, we can see that each of our model choices substantially improves its performance, with the biggest mAP improvements provided by our negative label expansion scheme. Each of the components of SCoNE also stacks additively, with our final model performing 2.4 mAP, 5.4 mR@15, and 5 F1@15 points over SB. Moreover, the components of SCoNE are \textit{model agnostic}. We verify that by enhancing our ResNet-Baseline with SCoNE, which also improves its mAP and mR@15 by 3.4 and 4.7 points.

Due to lack of space, several additional ablation experiments are included in the supplementary material.

\subsection{Qualitative examples}
Figure \ref{fig:exampls} shows qualitative results of our SB+SCoNE model, which clearly showcases its various strengths. Firstly, we clearly show a robust ability of the model to predict a variety of attribute types of different objects with good accuracy. Next, our object localizer shows remarkable ability to find the correct object of interest and ignore background and other distracting objects (in \ref{fig:exampls}a, the ground is not attended by the object localizer). Next, our multi-attention module often works to complement our object localizer by attending to relevant image regions that may be outside of the object region. For example, in figure \ref{fig:exampls}b, the activity \textit{skateboarding} is easier to predict if our model can look at the skateboard, but it is outside the \textit{person} region. Here, our multi-attention correctly learns to look at appropriate image regions that can help our model determine the \textit{person} has an attribute \textit{skateboarding}. More results can be found in the supplementary material.

\section{Discussion, Future works and Conclusion}

VAW is a first-of-its kind large-scale object attribute prediction dataset in the wild. We explored various challenges posed by the VAW dataset and also discussed efficacy of current models towards this task. Our SCoNE algorithm proposed several novel algorithmic improvements that have helped us improve performance in the VAW dataset compared to our strong baseline by 2.4 mAP and and 5.4 mR@15 points. Despite our results, there are several outstanding challenges remaining to be solved in VAW.

\noindent \textbf{Data imbalance:} Reweighting and resampling techniques have helped considerably improve the performance of tail categories in VAW dataset. However, even for our best model, mAP for tail categories still lags more than 25 points behind our head category. Similar to many vision and language problems \cite{kafle2019challenges}, this is one considerable challenge for future works in this space.

\noindent \textbf{Object-bias effect:} Using object label as input is crucial to obtain good results in VAW, but it may also introduce object-bias in predictions. Ideally, an algorithm should be able to make robust predictions for compositionally novel instance. While not in scope of current paper, this can be explored in detail by redistributing train-test split in compositionally novel patterns \cite{agrawal2018don,purushwalkam2019task,teney2020value}.

In conclusion, we believe that VAW can serve as an important benchmark not only for attribute prediction in the wild, but also as a generic test for long-tailed multi-label prediction task with limited labels, data imbalance, out-of-distribution testing and bias-related issues.

{\footnotesize \noindent\textbf{Acknowledgements}: This work was partially supported by DARPA SAIL-ON program (W911NF2020009) and gifts from Adobe collaboration support fund.}

{\small
\bibliographystyle{ieee_fullname}
\bibliography{egbib}
}

\clearpage
\appendix
\section*{Appendix}

\section{Example predictions from our SCoNE model}
\begin{figure*}
\centering
\includegraphics[width=1.0\linewidth]{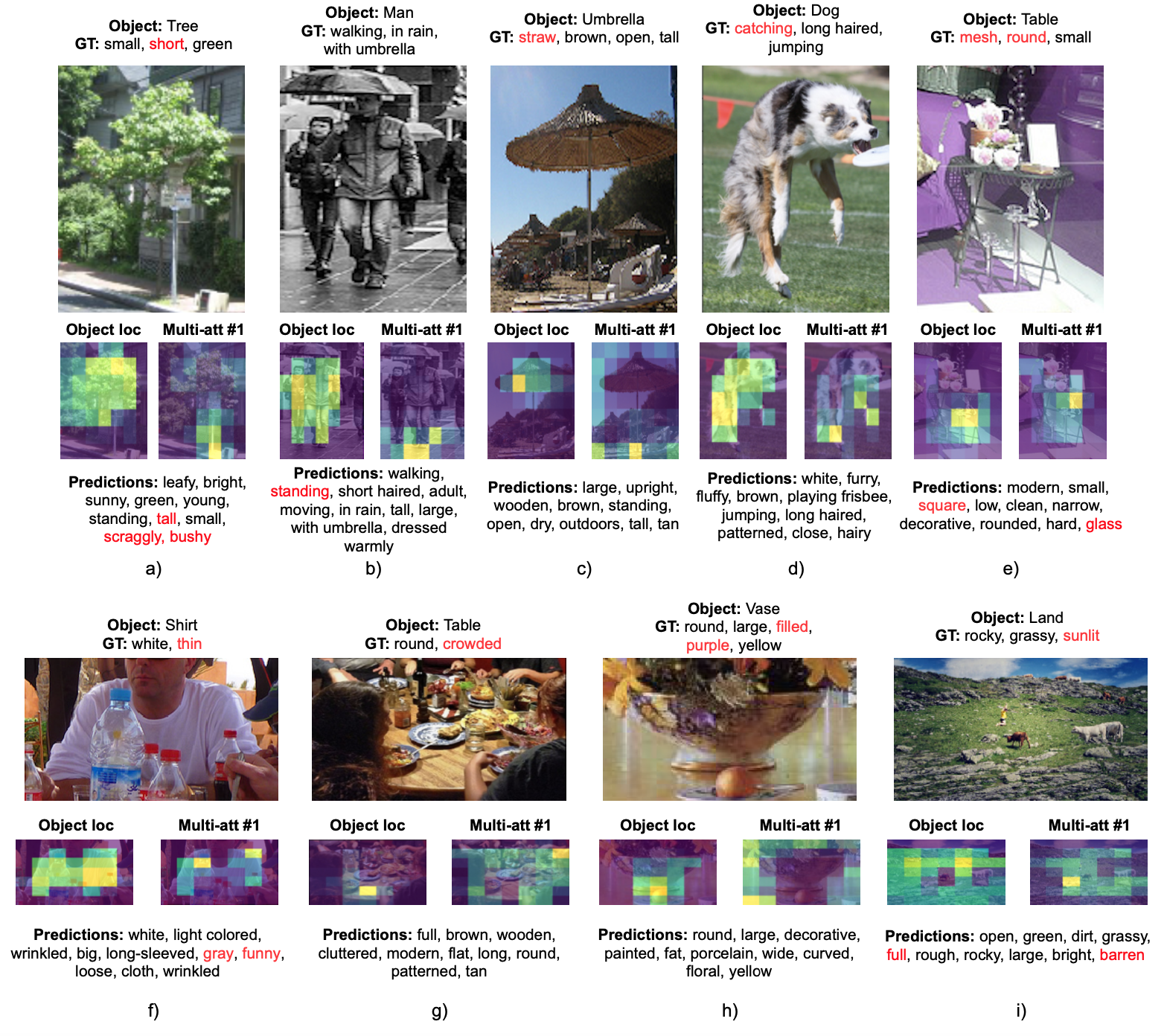}
    \caption{\textbf{Qualitative results.} Examples of predictions from SB+SCoNE. We show the object name and its ground truth positive attribute labels above the image. The object localized region, attention map \#1, and model top-10 predictions are shown below. Red text represents missed or incorrect predictions.}
\label{fig:examples}
\end{figure*}

Figure \ref{fig:examples} shows more qualitative results of our SCoNE algorithm together with the object localizer result as well as one (out of three total) attention maps from the multi-attention module. Throughout all examples, we can see that our model performs robustly for various attribute types. Our object localizer can correctly infer the object region and help alleviate the object occlusion problem, which is particularly a challenge for models that use global average pooling for feature aggregation. For example, in figure \ref{fig:examples}e, the object \textit{table} is partially occluded by a lot of clutter which can distract a model that relies on global average pooling. Here, our object localizer clearly isolates the parts of \textit{table}, making it easier to predict its attributes. However, object attributes are not always dependent on the foreground or surface features. Many attributes depend on the context, \eg., \textit{parked} vs. \textit{running car}. Therefore, we designed our multi-attention module to complement the object localizer by \textit{allowing} it to attend to image regions outside the object. This can be clearly seen in figure \ref{fig:examples}a, where attributes such as \textit{sunny} or \textit{bright} can be hard to infer by simply looking at the given object \textit{tree}. Our multi-attention module looks at the sunny spots on the pavement which can help our model infer the presence of \textit{sunny} and \textit{bright} attributes. However, the multi-attention module is also free to attend to regions in the object to further supplement object localizer's attention to specific parts of object. For example, in figure \ref{fig:examples}d, our multi-attention module attends to the hind legs of \textit{dog}, which supplements the object localizer's attention map and can provide additional information to help the model infer that the dog is \textit{jumping}.

\section{Additional details for the VAW dataset}
\begin{figure*}
\centering
\includegraphics[width=1.0\linewidth]{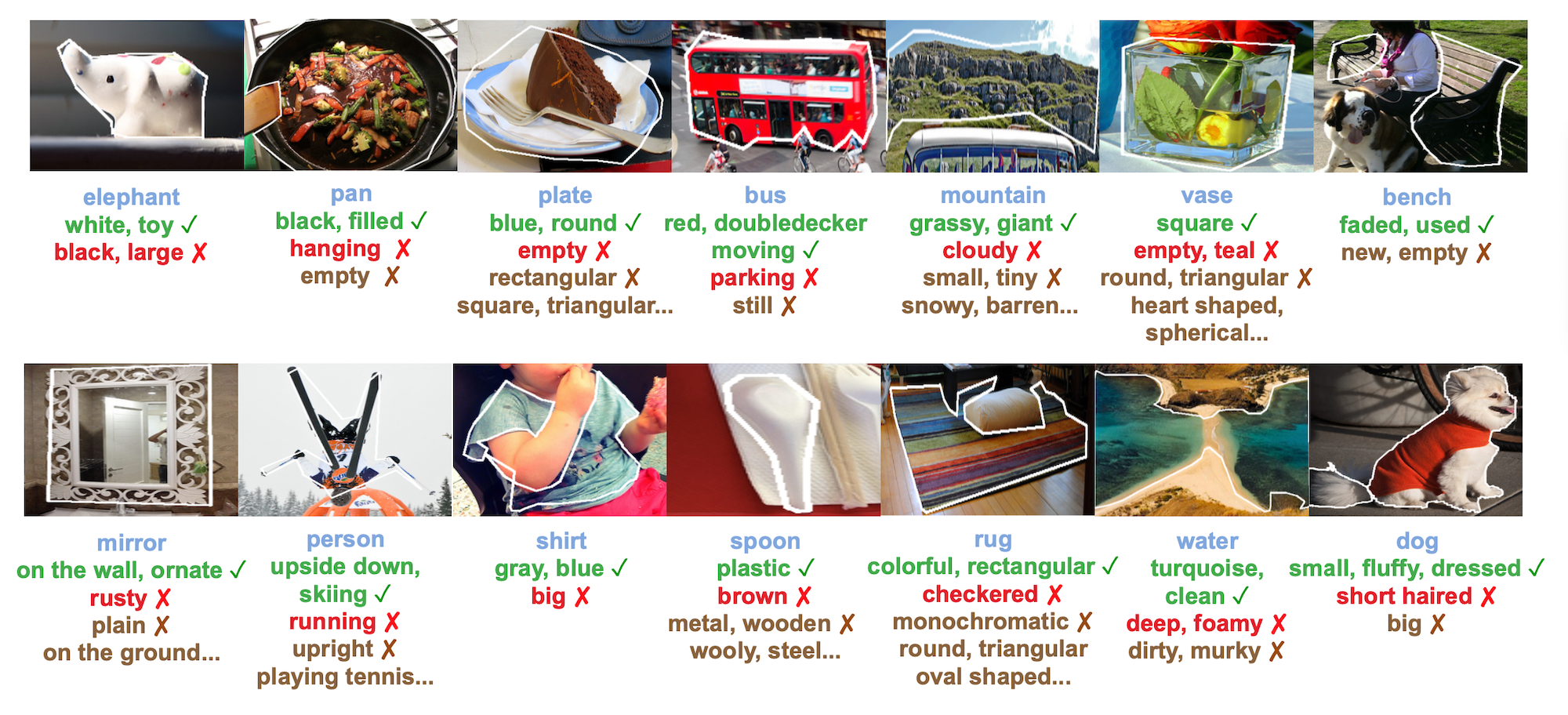}
    \caption{Examples of images and their annotations from the VAW dataset. \textcolor{CornflowerBlue}{Object names}, \textcolor{Green}{positive attributes}, \textcolor{red}{explicitly labeled negative attributes}, and \textcolor{brown}{negative labels from our negative label expansion} are shown in corresponding colors for each example.}
\label{fig:dataset_examples}
\end{figure*}

In figure \ref{fig:dataset_examples}, we show examples of images and their attribute annotations from the VAW dataset. The images show both positive and negative annotations from our dataset as well as a subset of the result of our negative label expansion scheme, which is a rule-based system derived on the premise of mutual exclusivity of certain attributes. For example, if an object is annotated with positive attribute \textit{empty}, the attribute \textit{filled} can be auto-annotated as a negative attribute for the same object.

\begin{figure*}
\centering
\includegraphics[width=1.0\linewidth]{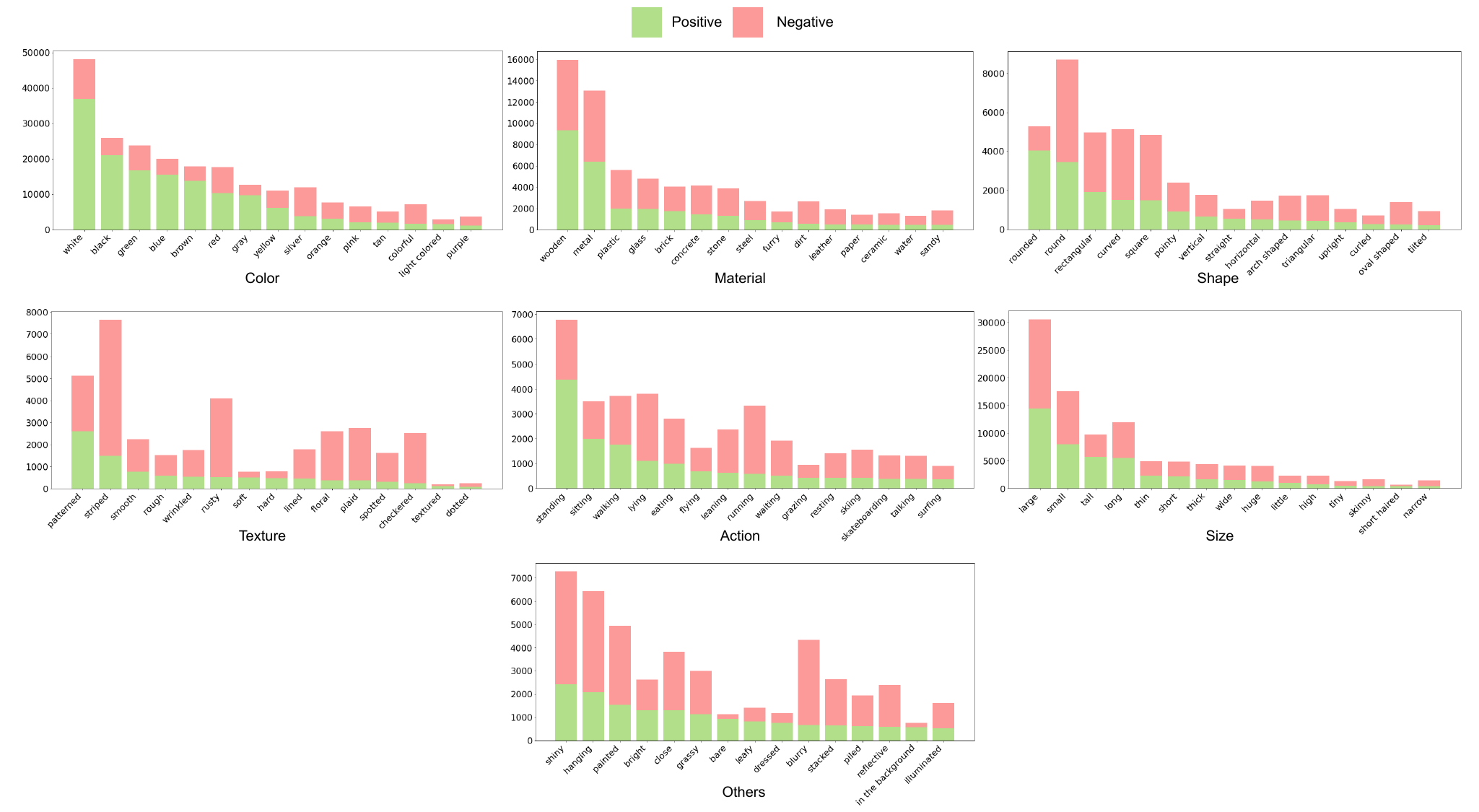} 
\caption{Distribution of positive and negative annotations for attributes in different categories. We show the top-15 attributes with the most number of positive annotations in each category sorted in descending order.}
\label{fig:dataset_distribution}
\end{figure*}

In figure \ref{fig:dataset_distribution}, we show the distribution of top-15 attributes in various attribute categories arranged in descending order according to the number of available positive annotations. The diagram clearly shows the long-tailed nature of our VAW dataset, with some categories showing highly skewed distributions (\textit{color, material}) and others have a more evenly balanced distribution (\textit{texture, others}). For example, in the \textit{material} category, the annotations for top-2 attributes (\textit{metal} and \textit{wooden}) consist of over 30.91\% of total number of annotations (41.4\% of positives and 23.5\% of negatives). Reassuringly, our strong baseline as well as SCoNE model works almost equally well for more balanced categories (\eg, \textit{texture}) as well as a skewed category (\eg, \textit{material}).

\section{Additional ablation studies}
\subsection{Study of different reweighting and resampling methods}

Our VAW dataset, by nature, has a large amount of data imbalance which is further exacerbated after our negative label expansion. Hence, we also studied various reweighting and resampling techniques to tackle this issue. Here, we show results for different methods that we considered. These methods include: (1) Class-aware sampling (CAS) \cite{shen2016relay}: fill classes in a training batch as uniform as possible; (2) Inverse frequency (IF) \cite{wang2017learning}: assign weight of each class to be inversely proportional to its frequency (applied with smoothing factor $\alpha = 0.1$); (3) Class-balanced (CB) \cite{cui2019class}: also a class-wise reweighting method similar to \cite{wang2017learning} but uses \textit{effective number} instead of actual number of positives $E_n = (1 - \beta^n)/(1 - \beta)$ with $\beta = 0.999$ and $n$ is the number of positives; (4) RW-BCE: our proposed reweighting scheme presented in section 4.4 in the main paper; ours is the only method among these that explicitly assigns different weight for the positive and negative label of every class (5) Repeat factor sampling (RFS) \cite{mahajan2018exploring,gupta2019lvis}: a resampling trick that oversamples images that contain the tail classes. 

The results are reported in table \ref{table:rr}. All these techniques are implemented on top of the ResNet-Baseline model and trained on the training data after negative label expansion.

CAS achieves low mAP score while still having decent mR@15 and mA. Because the VAW dataset is extremely imbalanced, applying CAS can lead to severe undersampling (or oversampling) of the head (or tail) classes. In addition, since CAS maintains a uniform distribution of classes in a training batch, no classes are dominant by the others as well as the negative examples do not dominate the positive examples. Hence, CAS still achieves good mean recall and mean accuracy.

IF and CB both assign higher weights for tail classes and achieve better results across all metrics over the baseline. Our formulation, RW-BCE, aims at (1) mitigating the oversuppression and rarity of negative examples and (2) highlighting the rare classes using the same weighting as in IF, hence, it achieves better results in all metrics over IF and CB. Finally, RFS is a resampling trick that does not rely on undersampling the head classes, thereby addressing one of the weaknesses of CAS, resulting in better performance in the VAW dataset. Because RFS is a sampling technique, it can be used in conjunction with any reweighting methods. Therefore, we use RW-BCE along with RFS (referred as RR) in our main paper whose results are better than the others across most metrics as shown in table \ref{table:rr}.

\begin{table} 
\resizebox{\columnwidth}{!}{
\begin{tabular}{lcccc}
    \toprule
    \textbf{Methods (+negative label expansion)} & \textbf{mAP} & \textbf{mR@15} & \textbf{mA} & \textbf{F1@15} \\
    \midrule
    ResNet-Baseline & 65.6 & 53.8 & 69.4 & 68.6 \\
    + Class-aware sampling (CAS) \cite{shen2016relay} & 63.5 & 56.6 & 70.2 & 65.4 \\
    + Class-balanced (CB) \cite{cui2019class} & 65.7 & 54.7 & 69.6 & 68.4 \\
    + Inverse frequency (IF) \cite{wang2017learning} & 65.8 & 54.8 & 69.6 & 68.4 \\
    + RW-BCE & \textbf{66.0} & 56.1 & 70.3 & \textbf{68.8} \\
    + Repeat factor sampling (RFS) \cite{mahajan2018exploring} \cite{gupta2019lvis} & 65.6 & 55.2 & 70.0 & 68.2 \\
    + RW-BCE + RFS & \textbf{66.0} & \textbf{57.0} & \textbf{70.6} & \textbf{68.8} \\
    \bottomrule
\end{tabular}
}
\\
\caption {Investigation of different reweighting and resampling methods.
}
\label{table:rr}
\vspace{0mm}
\end{table}

\subsection{Components of the Strong Baseline}
In the main paper, we presented ablations for our SB+SCoNE model, which is comprised of Strong Baseline, Negative label Expansion, and Supervised Contrastive Learning. However, Strong Baseline itself is comprised of many sub-components which extends the ResNet-Baseline: the object localizer, the multi-attention module, and the usage of low-level features. In this section, we will dissect how each of these components affect our Strong Baseline model.

We ablate our Strong Baseline model with each component and train on our training data after negative label expansion. We report results in table \ref{table:sb_ablation}. 

Removing each sub-component has a negative effect on the performance of the Strong Baseline model. For example, removing the use of low-level features not only lowers mAP in \textit{color} and \textit{material} attributes but it also lowers it for higher-level attributes (\eg, \textit{action}). This is likely due to the absence of clearly defined low- and high-level features, which forces a `single' feature to represent both low- and high-level features. This adversely affects the network's ability to learn high-level attributes (\eg., \textit{action}) as well as low-level (\textit{color, texture}), thus lowering performance for both.

Interestingly, removing the object localizer does not result in a drastically diminished performance. Visualizing the multi-attention output of our full model (Fig. \ref{fig:examples}) reveals that even without object mask supervision, the model is still able to differentiate between object and background/distractors with the multi-attention maps which are trained with weak supervision from the attribute labels. However, removing all components, which is devoid of any form of attention, 
severely hampers model performance across all categories. 

In general, all sub-components are necessary for our model to perform well across different attribute types. 

\begin{table*}[t]
\resizebox{\textwidth}{!}{
\centering
\begin{tabular}{@{}lccccccccccccc@{}}
    \toprule
    \multirow{2}{*}{\textbf{Methods (+Neg)}} &
    \multicolumn{4}{c}{\textbf{Overall}} & \multicolumn{3}{c}{\textbf{Class imbalance (mAP)}} & \multicolumn{5}{c}{\textbf{Attribute types (mAP)}} \\
    \cmidrule(r){2-5} \cmidrule(r){6-8} \cmidrule(r){9-14}
  & \textbf{mAP} & \textbf{mR@15} & \textbf{mA} & \textbf{F1@15} & \textbf{Head} & \textbf{Medium} & \textbf{Tail} & \textbf{Color} & \textbf{Material} & \textbf{Shape} & \textbf{Texture} & \textbf{Action} & \textbf{Others} \\
    \midrule
    Strong Baseline & 67.7 & 54.3 & 70.0 & 69.6 & 75.9 & 64.3 & 46.9 & 68.8 & 73.9 & 67.0 & 69.4 & 60.2 & 69.1 \\
    w/o Multi-attention (MA) & 67.4 & 53.5 & 69.7 & 69.7 & 75.9 & 63.8 & 46.4 & 67.8 & 74.7 & 66.9 & 68.5 & 58.0 & 69.0 \\
    w/o Low-level feature (LL) & 67.3 & 53.7 & 69.9 & 69.4 & 75.4 & 63.8 & 48.4 & 68.5 & 73.6 & 66.1 & 67.5 & 59.3 & 68.9 \\
    w/o Object localizer (OL) & 66.9 & 53.1 & 69.6 & 69.1 & 75.3 & 63.4 & 45.5 & 67.5 & 73.8 & 66.5 & 68.4 & 58.9 & 68.3 \\
    w/o OL, MA and LL & 65.6 & 53.8 & 69.4 & 68.6 & 74.8 & 62.3 & 43.2 & 67.3 & 73.3 & 66.3 & 67.7 & 56.0 & 67.4 \\
    \bottomrule
\end{tabular}
}
\\
\caption {Ablation study on the three components of the Strong Baseline model by removing each one. The last row also corresponds to the ResNet-Baseline model.
}
\label{table:sb_ablation}
\end{table*}

\subsection{Interaction of SupCon and attention module}
As presented in the main paper, using SupCon as a pretraining scheme can be at odds with the attention module. To investigate this issue, we compare between the following models that are trained on the training set after negative label expansion: (1) Strong Baseline, (2) Strong Baseline + SupCon pretraining, (3) Strong Baseline without multi-attention, (4) Strong Baseline without multi-attention + SupCon pretraining, and (5) Strong Baseline with SupCon joint training. The results are reported in table \ref{table:supcon_pretrain_and_joint}.

From the results, we can see that SupCon pretrained helps improve model performance for our strong baseline model variant without multi-attention. This clearly shows that SupCon is an effective technique. However, we can also see that the mAP score drops when using SupCon pretraining for the unmodified Strong Baseline model, which consists of multi-attention module. We conjecture that SupCon pretraining being incompatible with multi-attention is largely because SupCon pretraining uses global average pooling (GAP) for feature aggregation which encourages the feature extractor to ignore the surrounding context (the majority of attribute features lie on the object foreground which is in the image center), whereas the multi-attention module aims to detect features at different locations including the surrounding.

To alleviate this issue, we jointly train the supervised contrastive loss with our whole model. Results from table \ref{table:supcon_pretrain_and_joint} shows that SupCon joint training no longer experiences the above problem while improves almost all overall metrics. The benefit of SupCon joint training is even more evident in the tail classes.

Supervised contrastive learning is still a new learning approach with very few exploration in the community. We believe there will be better ways to incorporate supervised contrastive learning in a multi-label setting such as ours.

\begin{table*}[t]
\resizebox{\textwidth}{!}{
\centering
\begin{tabular}{@{}lccccccccccccc@{}}
    \toprule
    \multirow{2}{*}{\textbf{Methods (+Neg)}} &
    \multicolumn{4}{c}{\textbf{Overall}} & \multicolumn{3}{c}{\textbf{Class imbalance (mAP)}} & \multicolumn{5}{c}{\textbf{Attribute types (mAP)}} \\
    \cmidrule(r){2-5} \cmidrule(r){6-8} \cmidrule(r){9-14}
  & \textbf{mAP} & \textbf{mR@15} & \textbf{mA} & \textbf{OV-F1} & \textbf{Head} & \textbf{Medium} & \textbf{Tail} & \textbf{Color} & \textbf{Material} & \textbf{Shape} & \textbf{Texture} & \textbf{Action} & \textbf{Others} \\
    \midrule
    Strong Baseline & 67.7 & 54.3 & 70.0 & 69.6 & 75.9 & 64.3 & 46.9 & 68.8 & 73.9 & 67.0 & 69.4 & 60.2 & 69.1 \\
    + SupCon pretraining & 67.3 & 54.8 & 70.0 & 69.5 & 75.7 & 63.8 & 45.5 & 67.8 & 73.1 & 66.8 & 69.2 & 59.6 & 68.8 \\
    + SupCon joint training & 68.2 & 55.2 & 70.3 & 70.0 & 76.1 & 64.7 & 47.8 & 69.1 & 75.0 & 67.3 & 69.8 & 60.0 & 69.4 \\
    \midrule
    SB w/o Multi-attention & 67.4 & 53.5 & 69.7 & 69.7 & 75.9 & 63.8 & 46.4 & 67.8 & 74.7 & 66.9 & 68.5 & 58.0 & 69.0 \\
    + SupCon pretraining & 67.6 & 54.1 & 69.7 & 69.8 & 75.9 & 63.9 & 46.6 & 67.7 & 75.0 & 67.0 & 68.3 & 57.7 & 69.1 \\
    \bottomrule
\end{tabular}
}
\\
\caption {Experiments to show the incompatibility between SupCon pretraining and multi-attention used by both our Strong Baseline and SCoNE model. The top section shows results that accuracy decreases when using SupCon pretraining with multi-attention in Strong Baseline model, which can be alleviated by switching to jointly training. The bottom section shows that SupCon pretraining works well on its own when multi-attention is not being used.
}
\label{table:supcon_pretrain_and_joint}
\end{table*}

\section{Evaluation metrics}
In this section, we present details about the different evaluation metrics that we use. We have used mAP as our primary metric, since it describes the quality of the model to rank correct images higher than the incorrect ones for each attribute label. mR@15 is also important as it shows how well the model manages to output the ground truth positive attributes in its top 15 predictions in each image. In addition, mA and F1@15 can also be used to evaluate model performance in a different light.

\noindent \textbf{mAP:} similar to \cite{veit2017learning}, the mAP score is computed by taking the mean of the average precision of all $C$ classes
\begin{equation}
    mAP = \frac{1}{C} \sum_{c}AP_c,
\end{equation}
in which the average precision of each class is computed as
\begin{equation}
    AP_c = \frac{1}{P_c} \sum_{k=1}^{P_c} \text{Precision}(k, c) \cdot \text{rel}(k, c),
\end{equation}
where $P_c$ is the number of positive examples of class $c$, Precision$(k,c)$ is the precision of class $c$ when retrieving the best $k$ images, rel$(k,c)$ is the indicator function that returns 1 if class $c$ is a ground-truth positive annotation of the image at rank $k$. Note that due to VAW being partially labeled, we compute this metric only on the annotated data similar as in \cite{veit2017learning}. This evaluation scheme is also similar to what is used in \cite{gupta2019lvis}, where the authors introduce the definition of federated dataset. In this federated dataset setup, we only need for each label a positive and a negative set, then average precision for each label can be computed on these 2 sets.

\noindent \textbf{mA:} as in \cite{li2016richly,tang2019improving}, we compute the mean balanced accuracy (mA) to evaluate all models in a classification setting, using 0.5 as threshold between positive and negative prediction. Because our dataset is highly unbalanced between the number of positive and negative examples for some attributes, balanced accuracy is a good metric as it calculates separately the accuracy of positive and negative examples then take the average of them. In concrete, the mA score can be computed as follows
\begin{equation}
    mA = \frac{1}{C}\sum_c \Big( \frac{TP_c}{P_c} + \frac{TN_c}{N_c} \Big) / 2,
\end{equation}
where $C$ is the number of attribute classes, $P_c$ and $TP_c$ are the number of positive examples and true positive predictions of class $c$, and $N_c$ and $TN_c$ are defined similarly for the negative examples and predictions. Because mA uses threshold $0.5$, models that are not well-balanced between positive and negative prediction tend to receive low score.

\noindent \textbf{mR@15 and F1@15}: we follow \cite{gong2013deep} to compute the precision, recall and F1 score. For each image, we consider the top 15 predictions of the model as its positive predictions. These predictions are then compared with the ground-truth annotations to compute the metrics. Because VAW dataset is partially labeled, we only consider the predictions of labels that have been annotated, \ie if class $c$ is predicted on an image but that image is unannotated for class $c$, then the prediction is ignored. The overall precision and recall are computed as follows
\begin{align}
    \text{OV-Precision} = \frac{\sum_c TP_c}{\sum_c N_c^p}, \,\, \text{OV-Recall} = \frac{\sum_c TP_c}{\sum_c P_c},
\end{align}
where $TP_c$ is the number of true positives for attribute class $c$, $N_c^p$ is the number of positive predictions of class $c$, and $P_c$ is the number of ground truth positive examples of class $c$. With the same notations, the per-class precision and recall are computed as
\begin{equation}
    \text{PC-Precision} = \frac{1}{C}\sum_c \frac{TP_c}{N_c^p}, \,\, \text{PC-Recall} = \frac{1}{C}\sum_c \frac{TP_c}{P_c}.
\end{equation}

The F1 score is the harmonic mean of precision and recall, which is defined as
\begin{align}
    \text{OV-F1} &= \frac{2\times\text{OV-Precision}\times \text{OV-Recall}}{\text{OV-Precision} + \text{OV-Recall}}, \\
    \text{PC-F1} &= \frac{2\times\text{PC-Precision}\times \text{PC-Recall}}{\text{PC-Precision} + \text{PC-Recall}}.
\end{align}

In our paper, we report the per-class recall and overall F1 score, and we refer to them respectively as mR@15 and F1@15 throughout the text and our tables.

\section{Implementation details}
We use the ImageNet-pretrained \cite{deng2009imagenet} ResNet-50 \cite{he2016deep} as the feature extractor, and use the output feature maps from ResNet block 2 and 3 as low-level features. For the object name embedding, we use the pretrained GloVe \cite{pennington2014glove} 100-d word embeddings. We do not finetune these word embeddings during training as we want our model to generalize to unseen objects during test time.

We implement our model in PyTorch \cite{paszke2017automatic} and train using Adam optimizer with the default setting, batch size 64, weight decay of $1e-5$, an initial learning rate of $1e-5$ for the pretrained ResNet and $0.0007$ for the rest of the model. We train for 12 epochs and apply learning rate decay of 0.1 every time the mAP on the validation set stops improving for 2 epochs. We use image size 224x224 as input and basic image augmentations which include random cropping around object bounding box, random grayscale when an instance is not labeled with any color attributes, minor color jittering, and horizontal flipping. For each object bounding box in the dataset, we expand its width and height by $\mathrm{min}(w, h)\times 0.3$ to capture more context. For the hyperparameters, we set $\lambda_\text{fg} = 0.25, \lambda_\text{div} = 0.004$. In the multi-attention module, we select $D_\text{proj} = 128$ and use $M=3$ attention maps. Regarding reweighting and resampling, we use $t = 0.0006$ for RFS and $\alpha = 0.1$ for smoothing in the RW-BCE reweighting terms.

For SupCon pretraining, we pretrain on top of ImageNet-pretrained ResNet for 10 epochs with batch size 384 (768 views per batch), and initialize all matrices $A_c$ with the identity matrix. In the contrastive loss, we set temperature $\tau = 0.25$. We believe using a larger batch size will greatly benefit supervised contrastive pretraining as suggested by the authors \cite{khosla2020supervised}. For SupCon joint training with the other losses of the Strong Baseline model, we keep batch size as 64, we add $\lambda_\text{sup}\mathcal{L}_\text{sup}$ to the loss where $\lambda_\text{sup}=0.5$, and all other hyperparameters are the same as above.

\section{Additional details for negative label expansion}
We classify the attributes into types and construct their overlapping and exclusive relations using existing ontology from a related work \cite{Han2019Visual}, WordNet \cite{wordnet}, and the relation edges from ConceptNetAPI \cite{speer2016conceptnet}. Specifically:
\begin{itemize}
    \item Attribute categories: are automatically derived from WordNet hypernyms and ConceptNetAPI \textit{IsA} relation edge. These are then manually verified.
    \item Overlapping relations: from WordNet, we check if two attributes share the same synset (\eg, \textit{muddy} and \textit{dirty} share WordNet synset \textit{dirty.s.06}). From ConceptNetAPI, we use the following relation edges: \textit{Synonym, SimilarTo, DerivedFrom}.
    \item Exclusive relations: From WordNet, we use antonyms retrieved from the synsets' lemmas. From ConceptNetAPI, we use the following relations: \textit{Antonym, DistinctFrom}.
\end{itemize}

\section{Image search results from our SCoNE model}

\begin{figure*}
\centering
\includegraphics[width=0.93\linewidth]{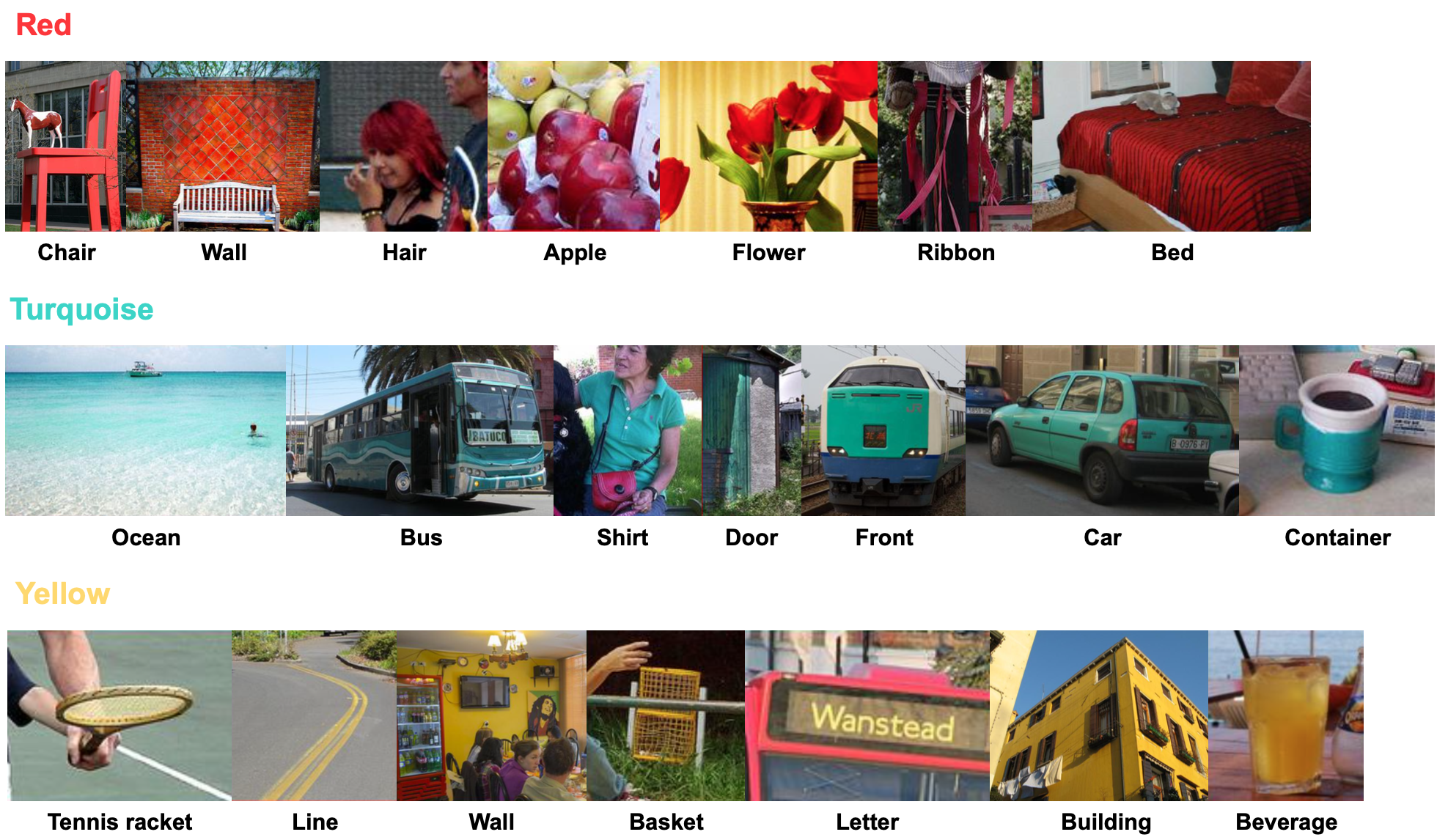}
    \caption{\textbf{Image search results.} We show the top retrieved images of SB+SCoNE when searching for some \textit{color} attributes.}
\label{fig:search_color}
\end{figure*}

\begin{figure*}
\centering
\includegraphics[width=0.93\linewidth]{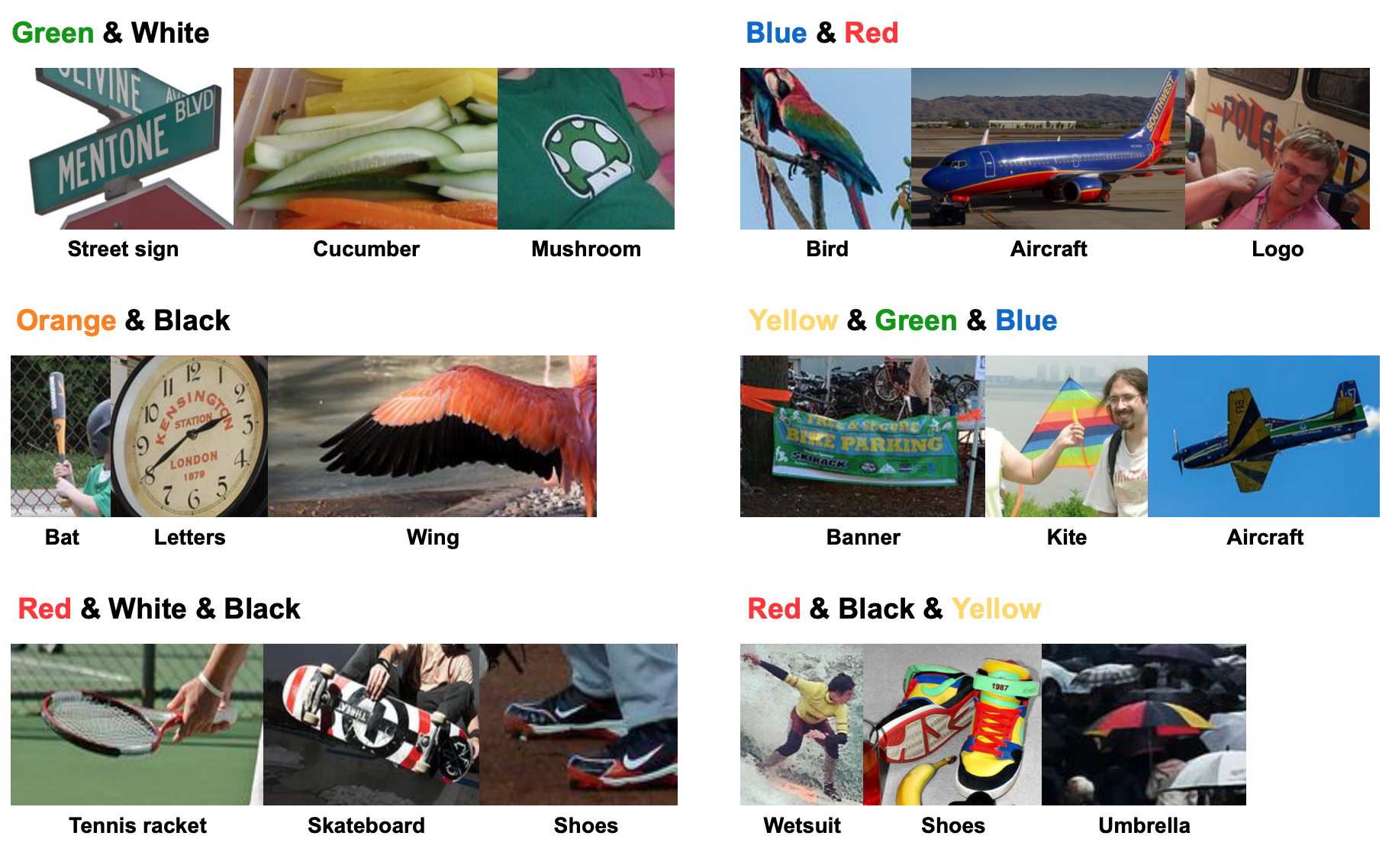}
    \caption{\textbf{Image search results.} We show the top retrieved images of SB+SCoNE when searching for images that exhibit multiple \textit{color} attributes.}
\label{fig:search_color2}
\end{figure*}

\begin{figure*}
\centering
\includegraphics[width=0.93\linewidth]{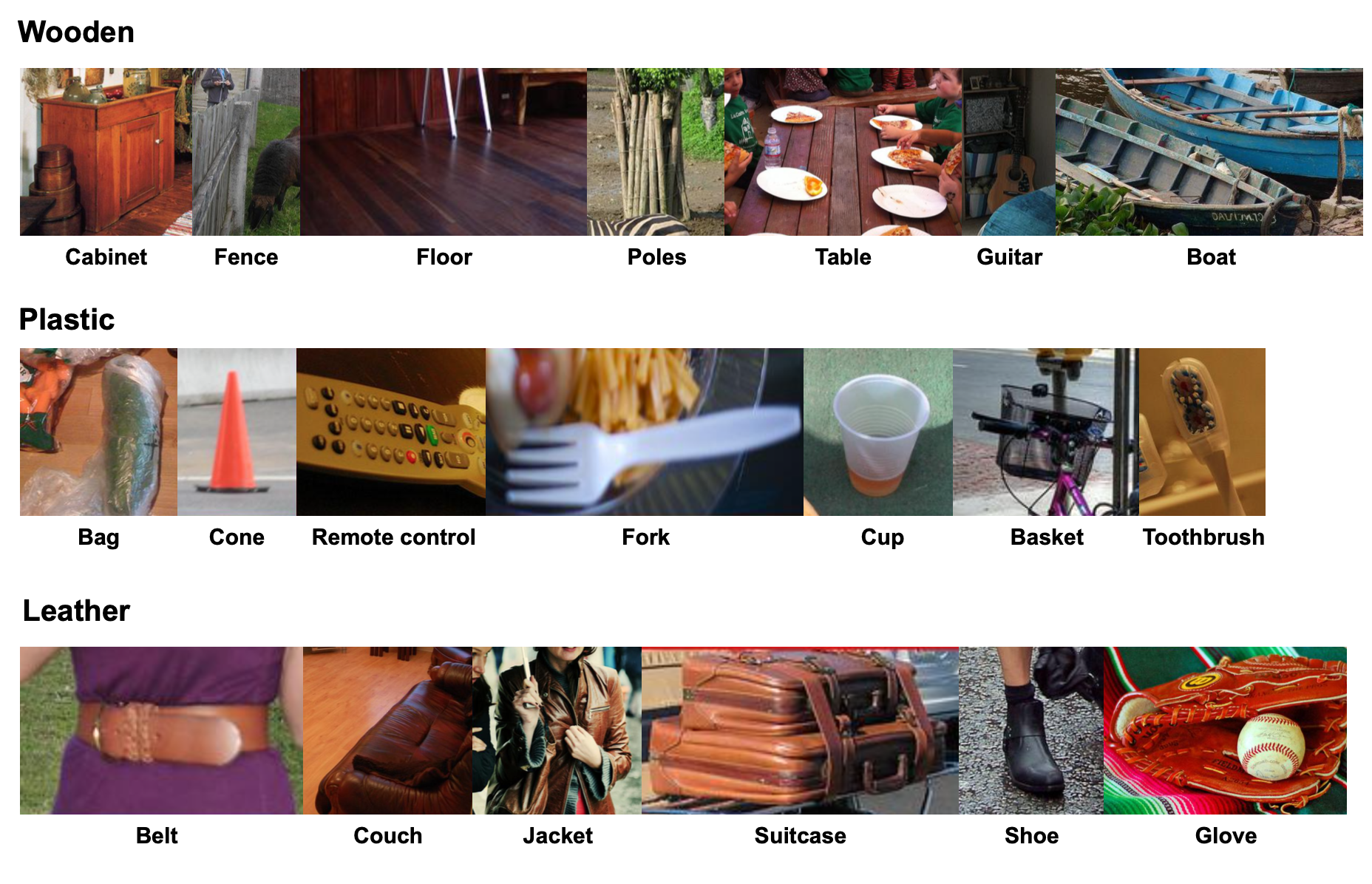}
    \caption{\textbf{Image search results.} We show the top retrieved images of SB+SCoNE when searching for some \textit{material} attributes.}
\label{fig:search_material}
\end{figure*}

\begin{figure*}
\centering
\includegraphics[width=0.93\linewidth]{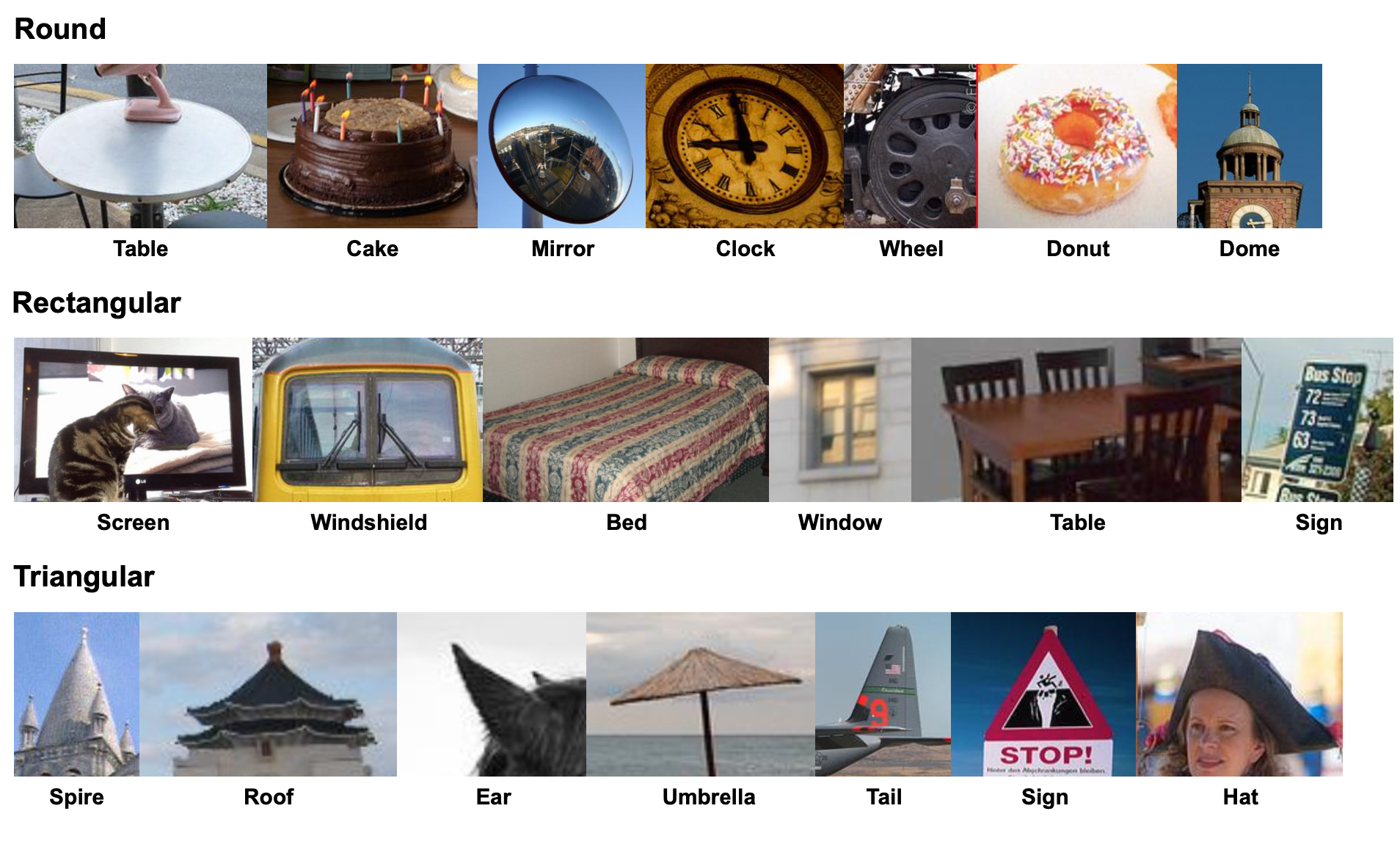}
    \caption{\textbf{Image search results.} We show the top retrieved images of SB+SCoNE when searching for some \textit{shape} attributes.}
\label{fig:search_shape}
\end{figure*}

\begin{figure*}
\centering
\includegraphics[width=0.9\linewidth]{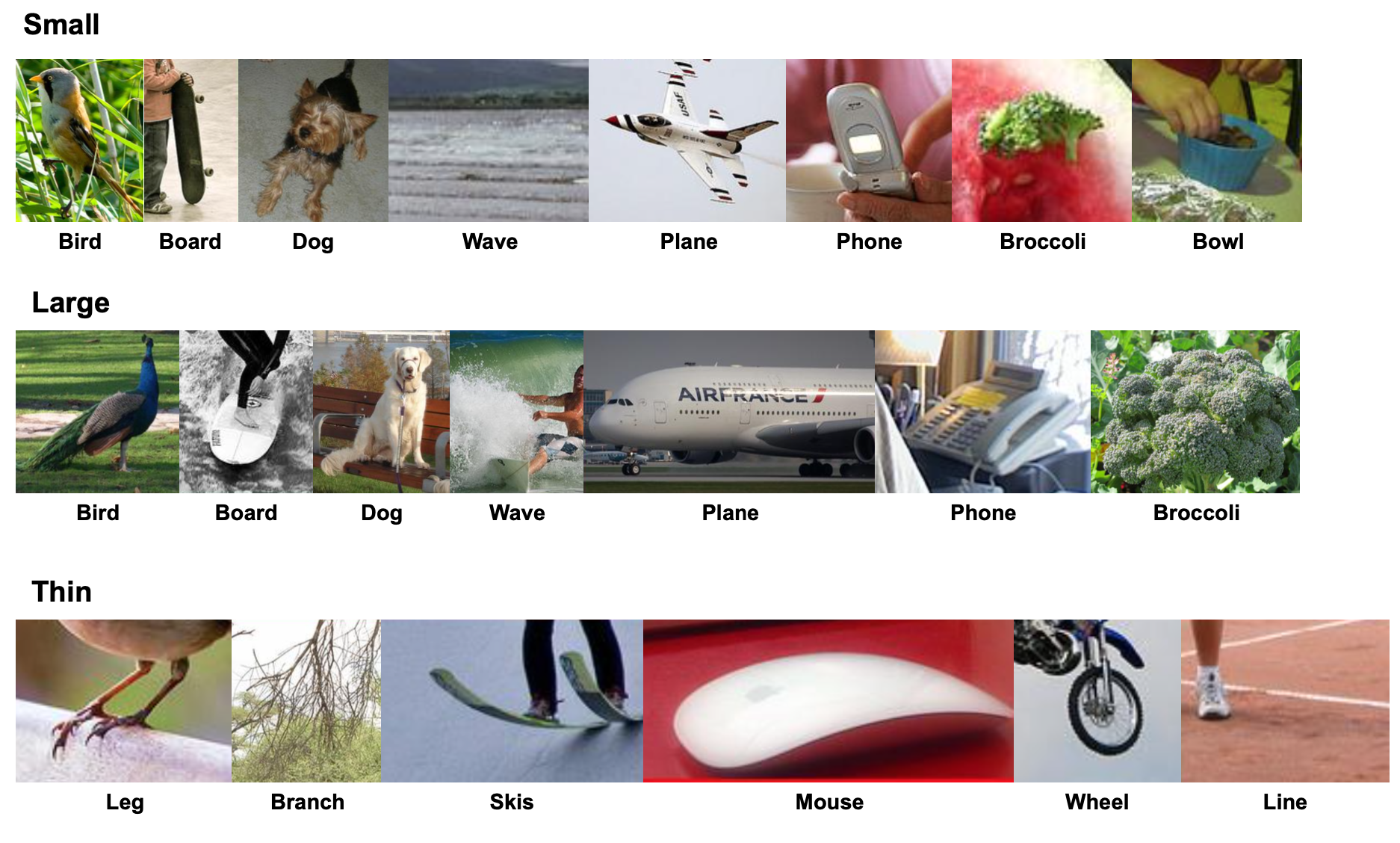}
    \caption{\textbf{Image search results.} We show the top retrieved images of SB+SCoNE when searching for some \textit{size} attributes. We deliberately show the same object categories between the 1st and 2nd row to show how our model is able to make distinction between a \textit{small bird} vs. \textit{large bird}, \textit{small plane} vs. \textit{large plane}, \etc.}
\label{fig:search_size}
\end{figure*}

We show from Figure \ref{fig:search_color} to Figure \ref{fig:search_size} our image search (ranking) results when searching for specific attributes. Our model is able to search for images that exhibit one to multiple attributes, as demonstrated in Figure \ref{fig:search_color2} where we search for multiple colors at a time. In addition, the results in Figure \ref{fig:search_size} also show that our model is able to differentiate between objects with different size (\eg, \textit{small} vs. \textit{large bird}, \textit{small} vs. \textit{large phone}).

\end{document}